# Emergent Goal-Directed Attention in Large Vision-Language Models

**Han Zhang**[a]
[a]University of Michigan Transportation Research Institute, Ann Arbor, MI, USA
Correspondence should be addressed to Han Zhang (hanzh@umich.edu)

**Human observers prioritize visual information according to task goals. Most computational models of naturalistic viewing are gaze-trained for free viewing, leaving open whether goal-directed attention can emerge in systems without gaze supervision. We tested two off-the-shelf vision-language models (VLMs), Qwen3-VL-32B-Thinking and Gemma-4-26B-A4B-it, on 4,887 naturalistic scenes under visual-search and free-viewing instructions. Model predictions were compared with human fixations on the same images under corresponding tasks. Both models aligned more closely with human fixations under matching goals than under mismatched goals. This crossover persisted in target-absent scenes, where alignment could not be explained by simple visual grounding, and appeared in decoder-layer readouts. Furthermore, model-thinking traces were grounded in target semantics during search and in visual prominence during free viewing. These findings show that general-purpose VLMs can generate human-aligned, goal-directed spatial priorities without gaze-specific training, informing theories of goal-directed attention and offering scalable tools for predicting where people look across tasks.**

## Introduction

A hallmark of human cognition is goal-directed attention, or the ability to flexibly control visual attention based on task goals. Consider a classic demonstration by Yarbus [1], in which observers viewed the same painting under different task instructions, such as estimating the ages of the people, remembering their clothes, or viewing the painting freely. Observers exhibited substantially different eye movement patterns depending on the task goal. This cognitive flexibility depends, in part, on observers' scene-semantic knowledge, such as a scene's overall meaning, the objects it typically contains, and where those objects are likely to appear [2–6]. For example, when searching for a keyboard in an office, observers draw on their knowledge of where keyboards are usually found to prioritize goal-relevant regions (e.g., the desk). By integrating scene-semantic knowledge with current task goals, observers can flexibly adjust their viewing behavior.

Given the importance of goal-directed attention, there has been substantial interest in developing computational models of human eye movements [7–14]. Early models focused on low-level image features, such as color, luminance, and orientation [8,12]. More recent deep-learning-based models have substantially improved predictive performance [9–11,13], suggesting that these systems capture important regularities in human fixation patterns. However, many of these models are developed to predict human fixations during free viewing and do not account for how attentional priorities change with task goals. Furthermore, because these models are typically trained directly on human fixation data, they cannot determine whether goal-directed attention can emerge in systems in a zero-shot manner without direct gaze training.

Large vision-language models (VLMs) offer an alternative approach to this question. These models are pretrained on vast amounts of multimodal data, including paired images and captions, web documents, knowledge bases, and code, and many are subsequently fine-tuned to follow natural-language instructions. Importantly, general-purpose VLMs are not typically trained to predict human fixations or fine-tuned on human gaze data. Through their broad training, however, VLMs may acquire scene-semantic knowledge analogous to that used by human observers, such as knowledge of a scene's overall meaning and where objects typically appear. More broadly, text-only large language models have shown human-like behavior on some cognitive tasks, including those involving decision-making, memory, and verbal reasoning [15–18]. This raises the question of whether general-purpose multimodal training and instruction-following capabilities might similarly enable VLMs to align with human goal-directed attention.

Here, we test two state-of-the-art, off-the-shelf VLMs, Qwen3-VL-32B-Thinking [19] and Gemma-4-26B-A4B-it [20], on thousands of naturalistic scenes and compare their output against human fixations on the same images. We prompted each model to point to the locations it would examine in a scene under two different task goals: searching for a specific target or freely viewing the scene for a memory test. Although VLMs do not literally search or freely view scenes, we adapted these prompts from the instructions given to human observers performing the corresponding tasks [21,22]. If VLMs exhibit human-like goal-directed behavior, their predictions should align more closely with human fixations when the model and human goals match than when they differ. We first tested this hypothesis using scenes in which the target was present and found greater human–model alignment when their goals matched than when they differed. As a stricter test, we replicated these effects in scenes in which the target was absent. Because these scenes contained no target, alignment with human search behavior provided stronger evidence that VLM predictions reflect scene-semantic knowledge rather than simple visual grounding. These goal-alignment effects also appeared in the models' internal decoder layers, suggesting that goal-related information is represented within the models' computational processes. As converging evidence, an analysis of Qwen's generated thinking traces showed that its reasoning was semantically closer to the target during visual search, but semantically closer to visual prominence during free viewing. Finally, both VLMs showed stronger fixation-order alignment in free viewing than in visual search. Overall, these results indicate an emergent ability in large VLMs to align with human goal-directed attention without direct gaze training.

# Results

To examine if VLMs acquired an emergent ability to align with human goal-directed attention, we asked VLMs to generate zero-shot predictions under two different task instructions on the same set of images. In the visual search condition, we asked VLMs to point out all locations they would examine before deciding whether the target was present. In the free-view condition, we instead asked VLMs to point out all locations they would examine to remember the scene (see Section "Prompts"). Human observers viewed the same images under similar instructions, with ten observers in each condition [21,22]. Because the stimuli are held identical, alignment in goal-directed attention would manifest as a crossover effect: model predictions under the search condition should match human search fixations more than free-view fixations. Conversely, model predictions under the free-view condition should match human free-view fixations more than search fixations.

## Target-present images

First, we examined a set of 2474 naturalistic scenes in which human observers either searched for an object present in the image or freely viewed the image for a memory test. Model predictions were derived for the same set of images, under both search and free-view instructions, and for both Qwen and Gemma. We constructed priority maps to indicate the spatial priority of human fixations and model-emitted points (see Figure 1 for an example).

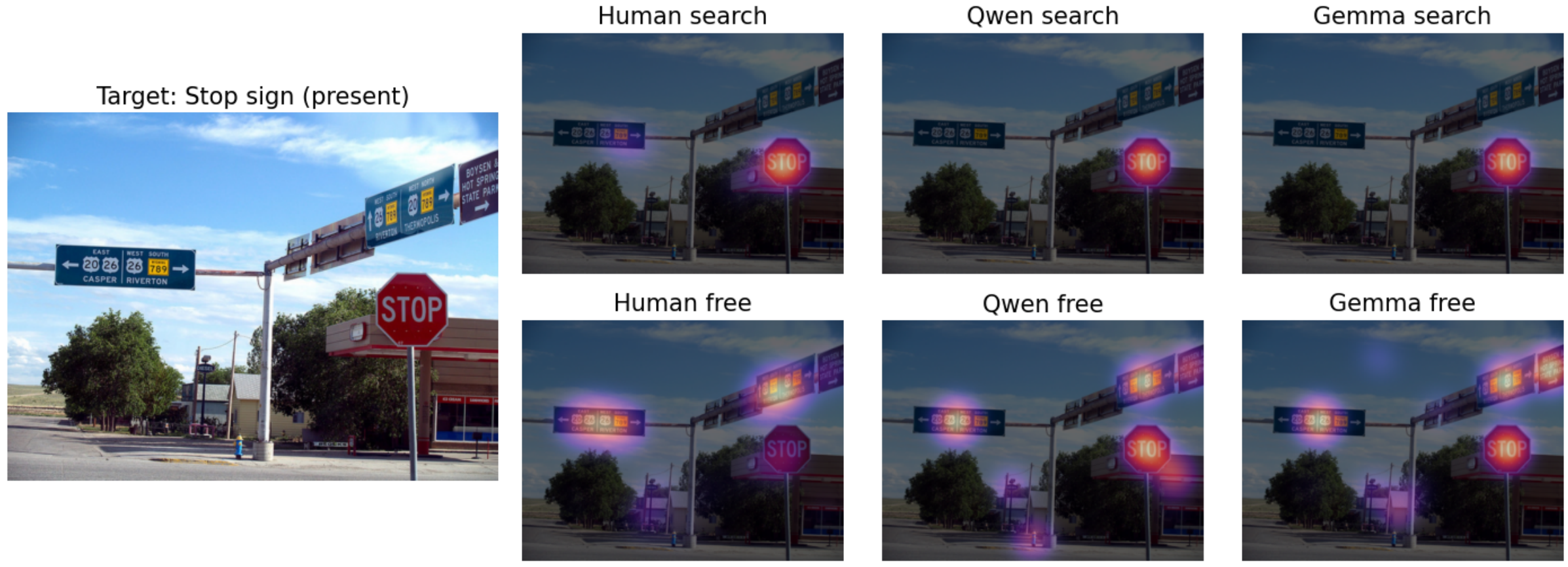

**Figure 1. Human and VLM priority maps for an example image.** In this case, the search target is a stop sign, which is present in the image.

We used Pearson correlation to quantify the spatial correspondence between human and model priority maps. The distributions of these correlations are shown in Figure 2, separately for model goal (search/free) and human goal (search/free). We conducted a 2*2 within-image repeated-measures ANOVA with model goal (search/free) and human goal (search/free) as factors. For Qwen, there was a main effect of model goal, $F(1, 2473) = 40.75$, $p < .001$, $\eta_G^2 = .003$, a main effect of human goal $F(1, 2473) = 5061.23$, $p < .001$, $\eta_G^2 = .274$, and importantly, a significant interaction between model goal and human goal, $F(1, 2473) = 12901.70$, $p < .001$, $\eta_G^2 = .48$. Planned comparisons show a crossover effect: Qwen search predictions aligned significantly better with human search fixations than with human free-view fixations (average $r_{search\text{-}search} = .73$, average $r_{search\text{-}free} = .15$; $t(2473) = 129.29$, $p < .001$, 95% Confidence Interval [0.578, 0.596], Cohen's $d = 2.91$). Conversely, Qwen free-view predictions aligned significantly better with human free-view fixations than with human search fixations (average $r_{free\text{-}free} = .48$, average $r_{free\text{-}search} = .35$; $t(2473) = 29.18$, $p < .001$, 95% CI [0.122, 0.139], $d = 0.77$).

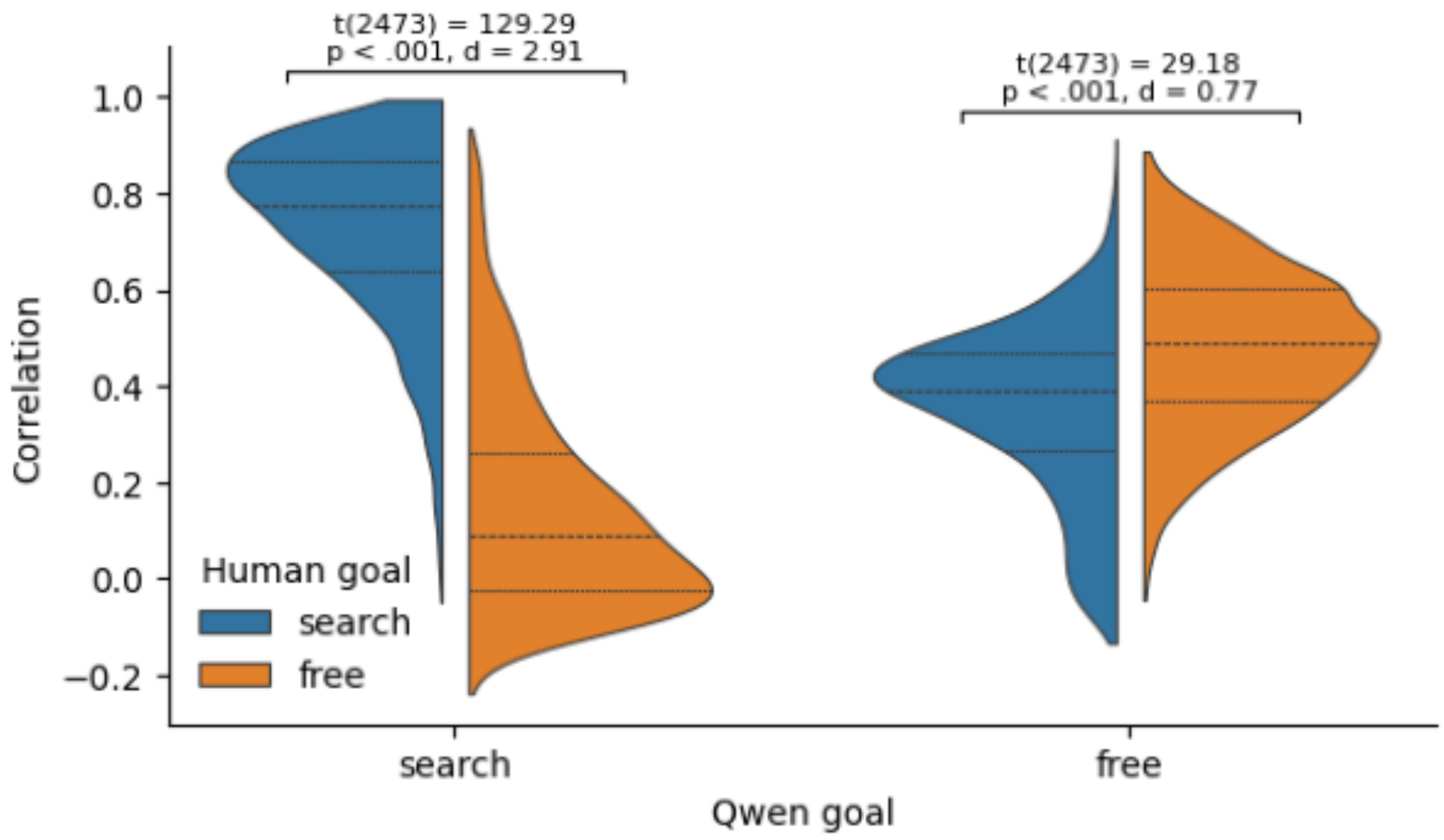

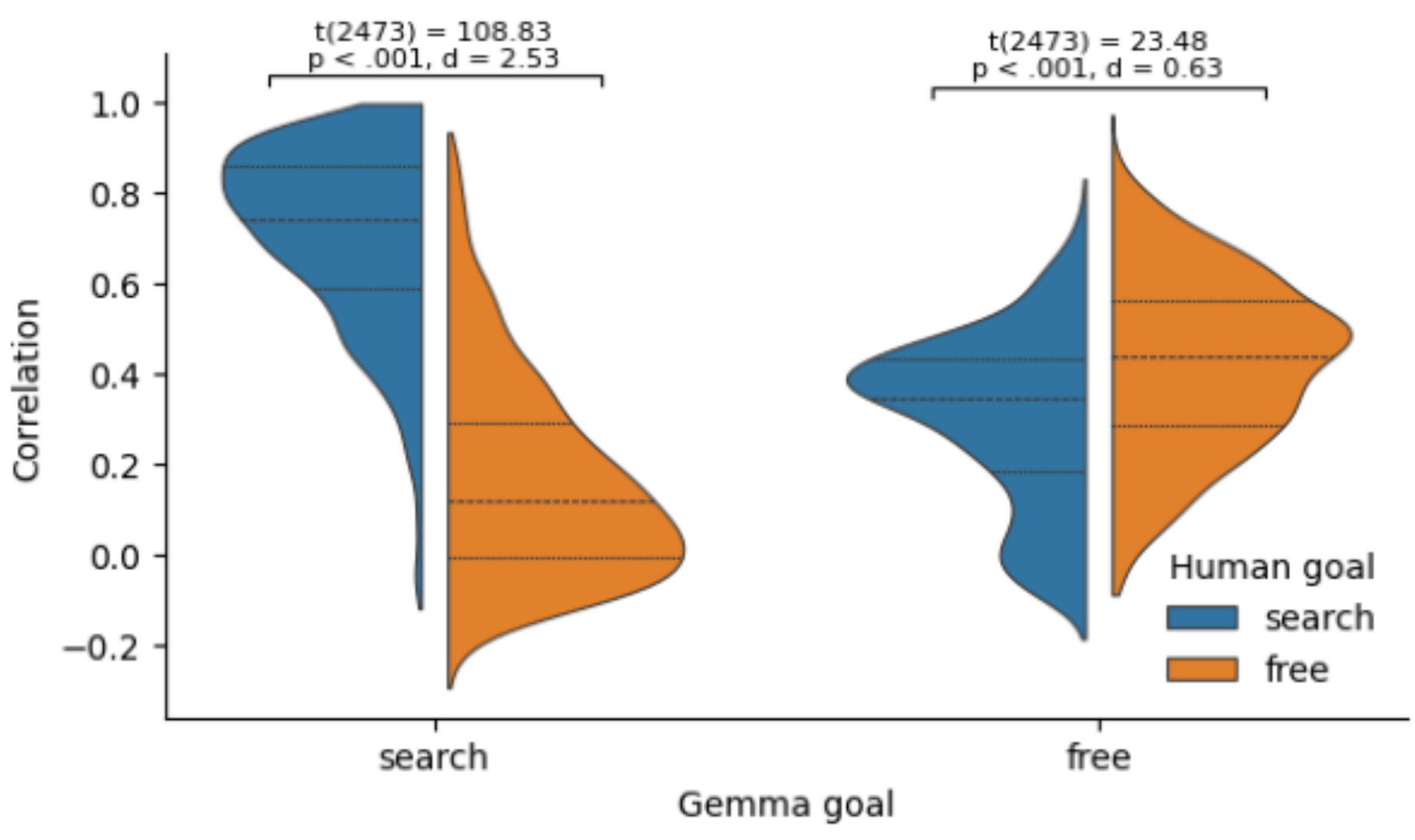

**Figure 2. Goal alignment between VLMs and human observers (target present).** The upper panel shows results for Qwen, and the bottom panel shows results for Gemma. The split violin plots show smoothed distributions of Pearson correlations between model and human priority maps on each image. The horizontal dashed lines inside the distributions denote quartiles. Alignment is indicated by the observation that model maps were more similar to human maps when the goal matched. The *t*-statistics and Cohen's *d*s are absolute values.

The results are similar for Gemma. There was a main effect of model goal, $F(1, 2473) = 369.36$, $p < .001$, $\eta_G{}^2 = .027$, and a main effect of human goal, $F(1, 2473) = 3210.62$, $p < .001$, $\eta_G{}^2 = .213$. Importantly, there was also a significant interaction between model goal and human goal, $F(1, 2473) = 9437.56$, $p < .001$, $\eta_G{}^2 = .400$. Specifically, Gemma search predictions aligned significantly better with human search fixations than with human free-view fixations (average $r_{search\text{-}search} = .70$, average $r_{search\text{-}free} = .16$; $t(2473) = 108.83$, $p < .001$, 95% CI [0.529, 0.548], $d = 2.53$). Similarly, Gemma free-view predictions aligned significantly better with human free-view fixations than with human search fixations (average $r_{free\text{-}free} = .43$, average $r_{free\text{-}search} = .31$; $t(2473) = 23.48$, $p < .001$, 95% CI [0.109, 0.129], $d = 0.63$).

On target-present images, both humans and VLMs tend to fixate or point at the target directly (see Figure 1). Thus, the high model-human alignment under the search condition could simply reflect VLM's visual grounding ability, without any knowledge of where objects tend to appear given the scene context. Target-absent images provide a stricter test. VLMs were not trained to predict where an absent target should be or fitted to human fixations. As such, any alignment with human fixations must instead draw on acquired knowledge of scene semantics.

## Target-absent images

We examined a set of 2413 naturalistic scenes in which the target is absent under the search condition. Figure 3 shows human and model priority maps for an example scene, in which the target under the search condition is a bottle.

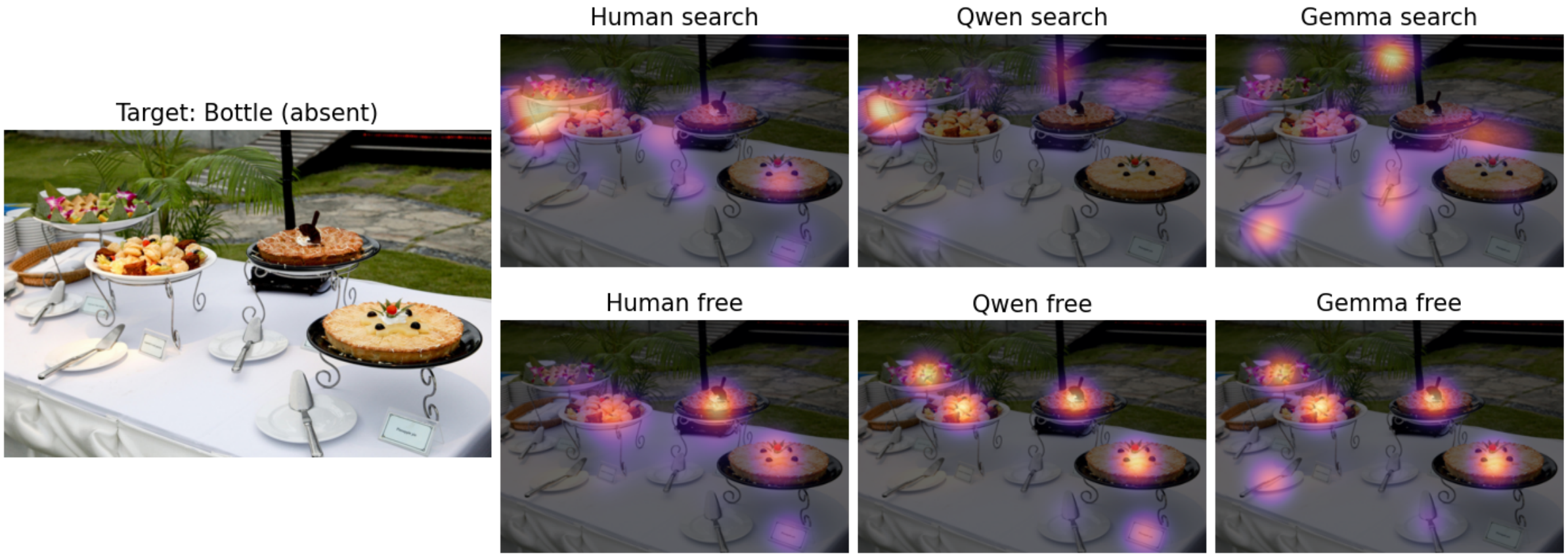

**Figure 3. Human and VLM priority maps for an example image.** In this case, the search target is a bottle, which is absent from the image.

Figure 4 shows the distributions of correlations for each combination of human and model goals. We again conducted a 2*2 repeated-measures ANOVA with model goal (search/free) and human goal (search/free) as factors. For Qwen, there was a main effect of model goal, $F(1, 2412) = 311.35$, $p < .001$, $\eta_G^2 = .045$, and a main effect of human goal, $F(1, 2412) = 470.52$, $p < .001$, $\eta_G^2 = .020$. Importantly, the interaction between model goal and human goal was also significant, $F(1, 2412) = 3873.89$, $p < .001$, $\eta_G^2 = .174$. Planned comparisons show that Qwen search predictions aligned significantly better with human search fixations than with human free-view fixations (average $r_{search\text{-}search} = .45$, average $r_{search\text{-}free} = .19$, $t(2412) = 56.67$, $p < .001$, 95% CI [0.254, 0.273], $d = 1.02$). Conversely, Qwen free-view predictions aligned significantly better with human free-view fixations than with human search fixations (average $r_{free\text{-}free} = .48$, average $r_{free\text{-}search} = .35$, $t(2412) = 34.62$, $p < .001$, 95% CI [0.130, 0.146], $d = 0.81$).

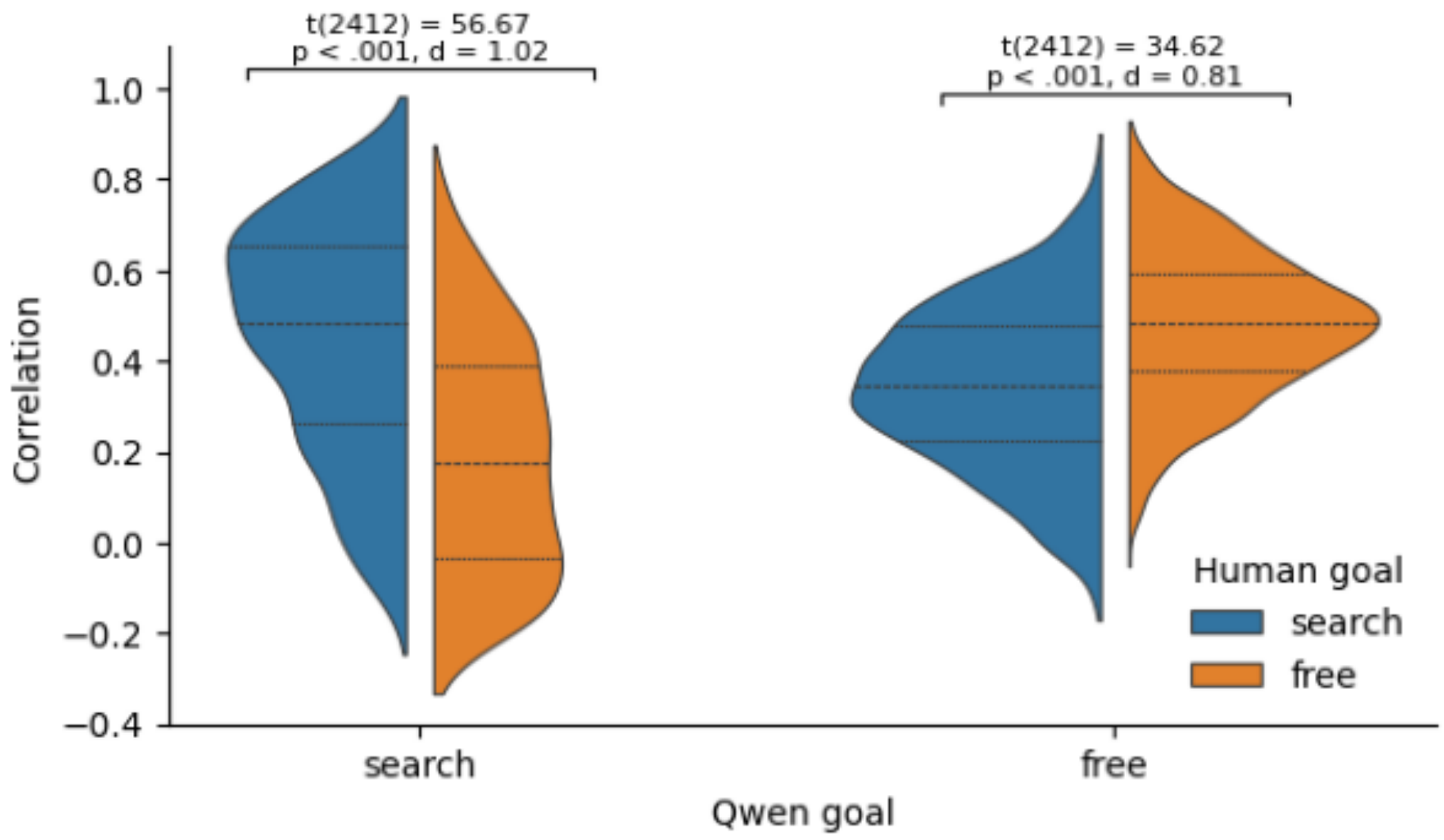

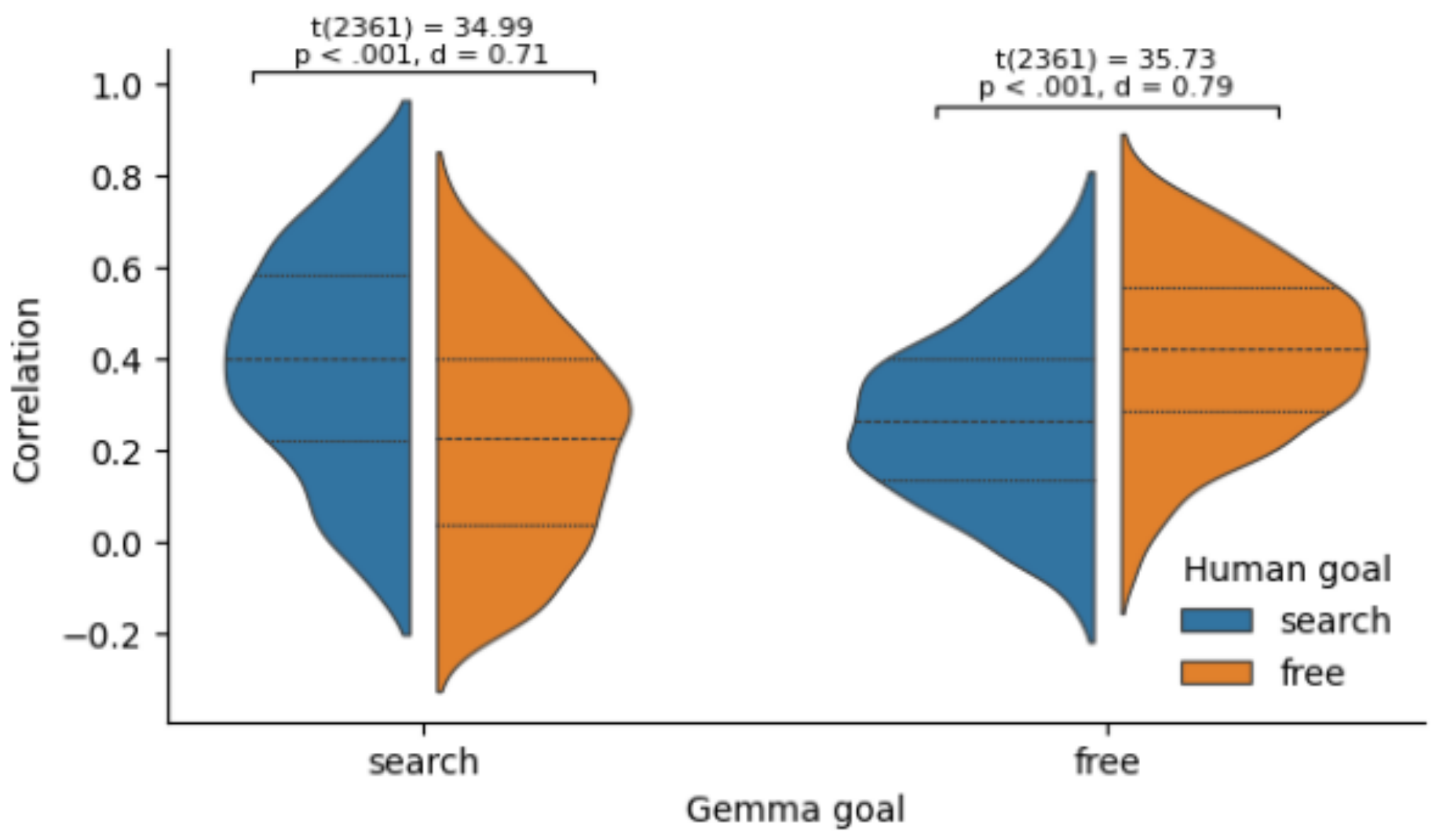

**Figure 4. Goal alignment between VLMs and human observers (target absent).** The upper panel shows results for Qwen, and the bottom panel shows results for Gemma. The split violin plots show smoothed distributions of Pearson correlations between model and human priority maps on each image. The horizontal dashed lines inside the distributions denote quartiles. Alignment is indicated by the observation that model maps were more similar to human maps when the goal matched. The *t*-statistics and Cohen's *d*s are absolute values.

Gemma's results also show a crossover pattern. There was a main effect of model goal, $F(1, 2361) = 38.60$, $p < .001$, $\eta_G^2 = .006$, a main effect of human goal, $F(1, 2361) = 15.73$, $p < .001$, $\eta_G^2 < .001$, and a significant interaction between model goal and human goal, $F(1, 2361) = 2367.97$, $p < .001$, $\eta_G^2 = .120$. Planned comparisons show that Gemma search predictions aligned significantly better with human search fixations than with human free-view fixations (average $r_{search\text{-}search} = .40$, average $r_{search\text{-}free} = .22$, $t(2361) = 34.99$, $p < .001$, 95% CI [0.163, 0.182], $d = 0.71$). Conversely, Gemma's free-view predictions aligned significantly better with human free-view fixations than with human search fixations (average $r_{free\text{-}free} = .42$, average $r_{free\text{-}search} = .27$, $t(2361) = 35.73$, $p < .001$, 95% CI [0.139, 0.155], $d = 0.79$).

## Direct readout from decoder layers

The results so far indicate that VLMs generate zero-shot predictions that align with human goal-directed fixations. To test whether such goal alignment is intrinsic to the model's computational processes, we directly extracted attention weights from Qwen's and Gemma's decoder layers and constructed priority maps based on these extracted weights. The results are presented in the Supplementary Materials. Overall, priority maps constructed from decoder layers show weaker correspondence to human priority maps than final model answers. However, the critical crossover effect was still replicated for both target-present and target-absent images, and for both Qwen and Gemma.

## Scene-level alignment

To further understand how VLMs align their predictions at the scene level, we created contrast maps by subtracting free-view priority maps from search priority maps for each scene. These contrast maps show which areas in the scene were prioritized and deprioritized by humans and models as task instructions changed (see Figure 5 for an example). A fidelity metric can be computed as the correlation between a model contrast map and a human contrast map. High fidelity, therefore, indicates that VLMs reorganize predictions to closely track human spatial priority at the level of each scene.

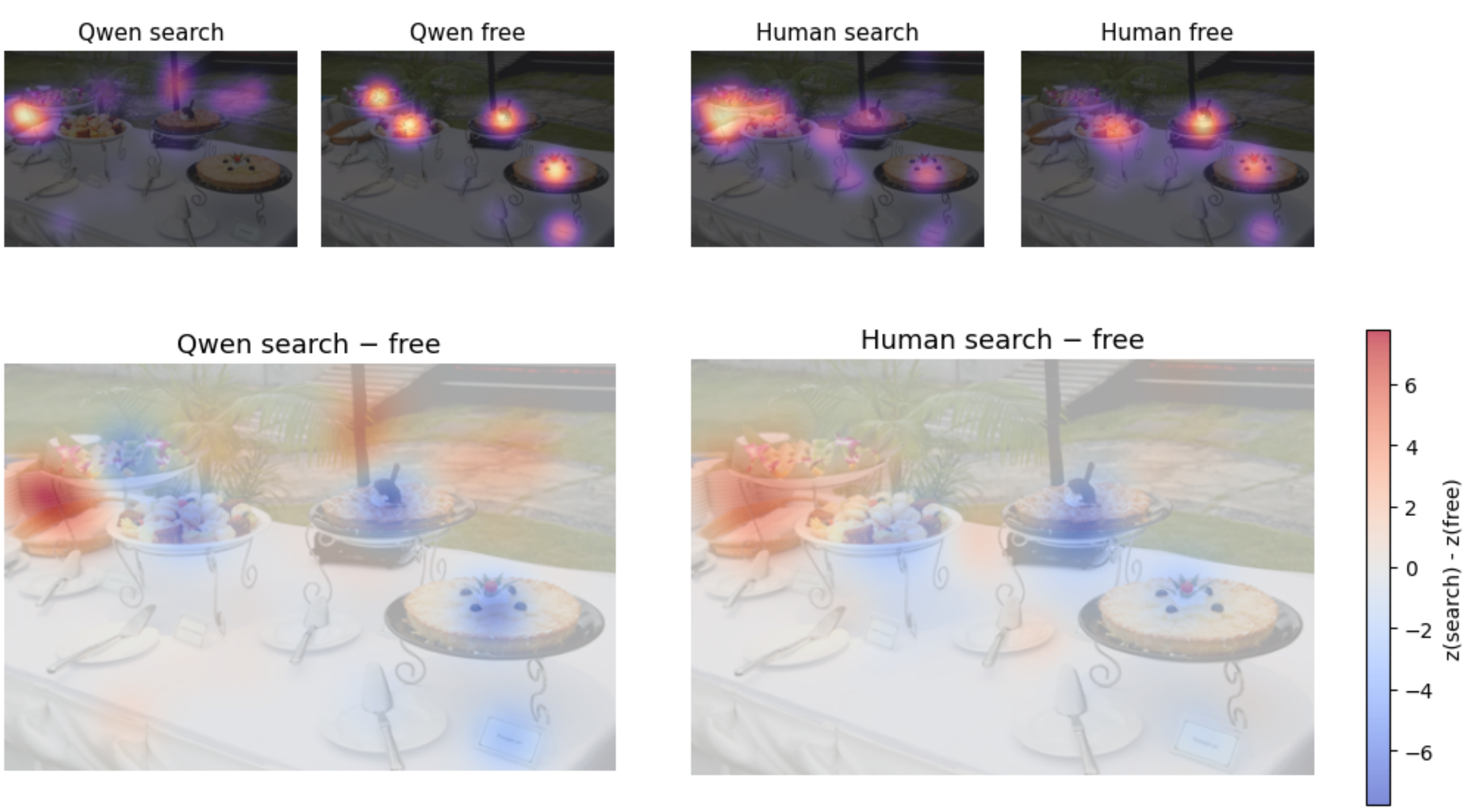

**Figure 5. A contrast map between search and free view to isolate areas prioritized in each condition.** The left side shows the contrast map for Qwen, and the right side shows the contrast map for human observers. A "fidelity" measure, defined as the correlation between these two contrast maps, indicates whether the model prioritizes and deprioritizes the same regions as humans when the goal changes.

Figure 6 shows fidelity distributions for target-present and target-absent images for both models. For target-present images, Qwen shows a mean fidelity of 0.530 (one-sample $t(2473) = 166.52$, $p < .001$, 95% CI [0.524, 0.537], $d = 3.35$), with over 99% of images showing positive fidelity. Gemma shows a mean fidelity of 0.487 (one-sample $t(2473) = 134.17$, $p < .001$, 95% CI [0.480, 0.494], $d = 2.70$), with 99% of images showing positive fidelity. For target-absent images, Qwen shows a mean fidelity of 0.310 (one-sample $t(2412) = 73.29$, $p < .001$, 95% CI [0.301, 0.318], $d = 1.49$), with 92% of images showing positive fidelity. Gemma shows a mean fidelity of 0.249 (one-sample $t(2361) = 55.11$, $p < .001$, 95% CI [0.240, 0.258], $d = 1.13$), with 87% of images showing positive fidelity. A supplemental analysis shows that the mean fidelity value was positive for every target object category (see Supplementary Materials). Overall, these results indicate that VLMs overwhelmingly align scene-level predictions in the same direction as human observers.

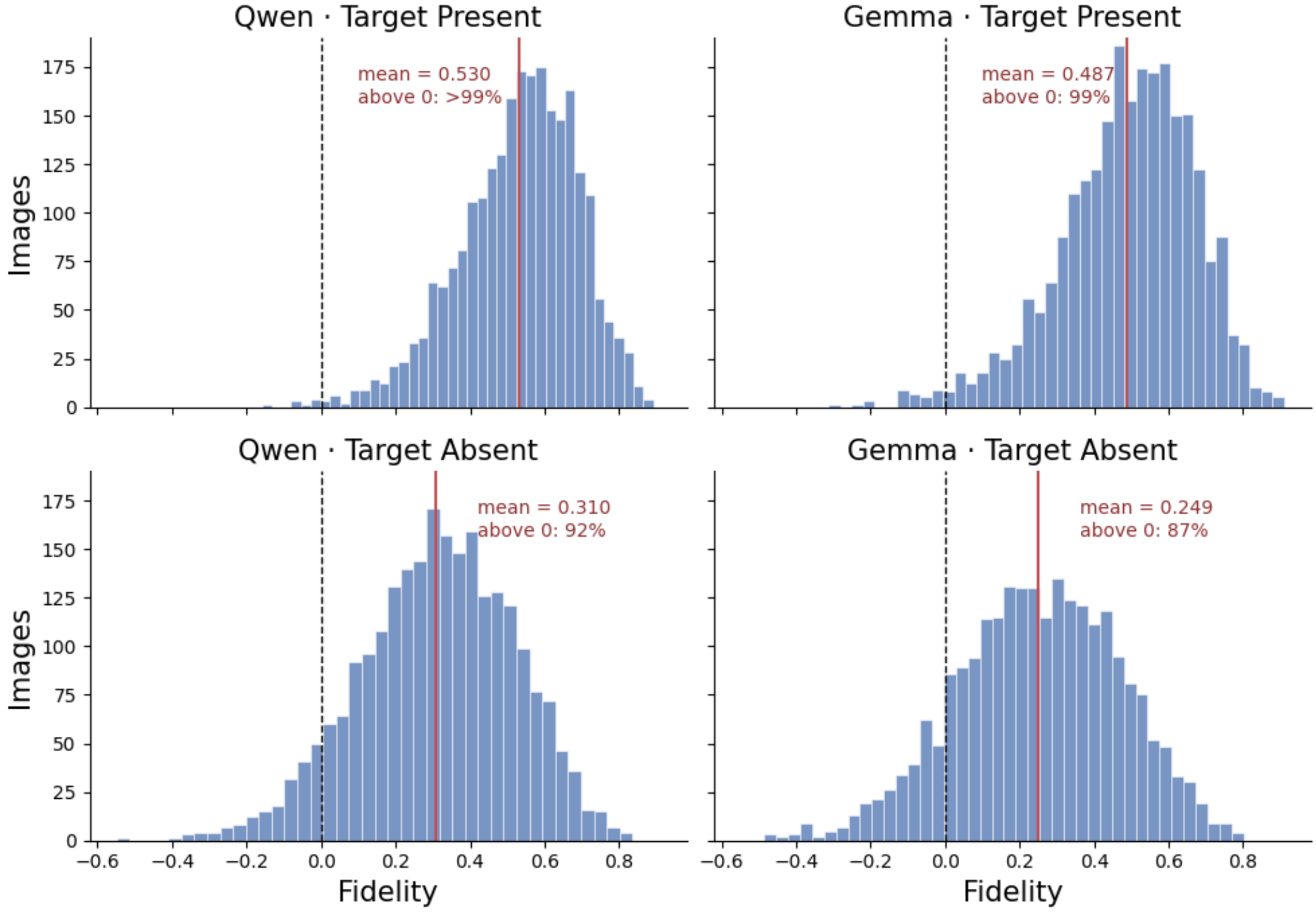

**Figure 6. A histogram of the fidelity measure for each model and target presence condition.** In all cases, most of the distribution is above zero, indicating that the VLMs in general prioritize and deprioritize scene regions in the same direction as humans when the goal changes.

## Alignment in model thinking traces

As a thinking model, Qwen additionally emits step-by-step, human-readable thinking traces before producing a final answer. Leveraging this feature, we asked whether the goal-directed behavior appears in the model's reasoning. Specifically, we asked whether search reasoning is organized around looking for regions that are semantically close to the target, while free-view reasoning is organized around looking for regions that are visually prominent. For example, when searching for an absent bottle in this scene (Figure 3), the reasoning appears to be rooted in where a bottle would typically be (e.g., "*Let's think of common places where a [target] might be: near the table, under the table, behind the dishes.*"). Under the free-view instruction for the same scene, the reasoning appears to be rooted in visually prominent content (e.g., "*The pineapple pie ... is very eye-catching.*").

To formally test this semantic dissociation, we constructed two semantic anchors. For the search condition, the anchor is simply the target word. For the free-view condition, the semantic anchor is a set of seven terms denoting visual prominence: *salient*, *conspicuous*, *prominent*, *striking*, *noticeable*, *dominant*, and *stands out*. We extracted Qwen's thinking traces and converted the cleaned traces into embeddings using a lightweight sentence transformer model [23]. Importantly, target words were removed from the cleaned sentences to avoid artificially inflating semantic alignment. The two semantic anchors were also embedded, with the seven terms in the free-view condition averaged to obtain a single embedding vector. Cosine similarity was computed between each thinking vector and each anchor vector. These similarity scores were then averaged at the scene level to indicate how well the model's thinking aligns with the target or visual prominence under search and free-view conditions.

Figure 7 shows distributions of semantic similarity between Qwen's thinking and the two semantic anchors. We conducted a 2*2 repeated-measures ANOVA with model goal (search/free-view) and semantic content (target/visual prominence) as factors. The dependent variable was the mean cosine similarity per image. The main effect of model goal was significant, $F(1, 2415) = 532.81$, $p < .001$, $\eta_G^2 = .018$. The main effect of semantic content was also significant, $F(1, 2415) = 383.73$, $p < .001$, $\eta_G^2 = .067$. Importantly, there was a significant interaction between model goal and semantic content, $F(1, 2415) = 7175.84$, $p < .001$, $\eta_G^2 = .25$. Planned comparisons show that search reasoning was more target-aligned than free-view reasoning, $t(2415) = 38.39$, $p < .001$, 95% CI [0.033, 0.037], $d = 0.67$. On the other hand, free-view reasoning was more prominence-aligned than search reasoning, $t(2415) = 127.55$, $p < .001$, 95% CI [0.055, 0.057], $d = 2.79$. A supplemental analysis shows that this dissociation was observed across all target object categories (see Supplementary Materials).

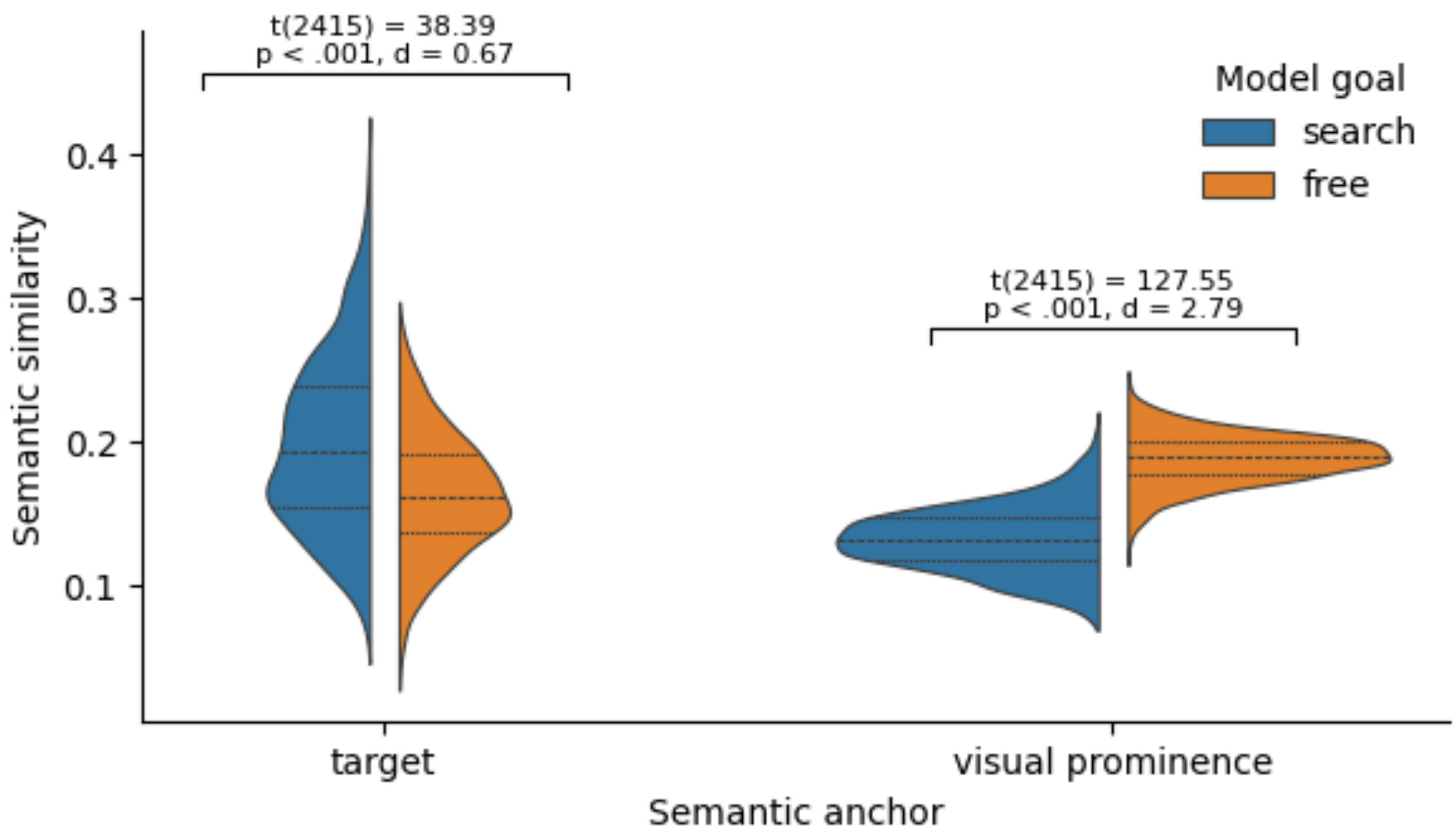

**Figure 7. Semantic similarity between Qwen's thinking traces and semantic anchors under different task goals.** The split violin plots show smoothed distributions of average cosine similarity scores between Qwen's thinking traces for each image and the two semantic anchors. The horizontal dashed lines inside the distributions denote quartiles. Alignment is indicated by the observation that the model changed its reasoning to align more closely with one semantic anchor over another, depending on the goal. The *t*-statistics and Cohen's *d*s are absolute values.

## Alignment in fixation order

Finally, we examined whether VLM predictions also sequentially align with human fixations beyond spatial alignment. We computed scanpath similarity between model-emitted points and human fixation order using a distance-based string-edit distance metric [24]. To isolate fixation-order alignment, we constructed a baseline in which the same scanpath similarity metric was computed using shuffled model-emitted points that randomized the order while preserving the overall spatial layout. Fixation-order alignment is, therefore, the scanpath similarity between model predictions and human fixations beyond the shuffled floor.

Figure 8 shows distributions of scanpath similarity results relative to the shuffled floor. We conducted a 2*2 repeated-measures ANOVA with model type (Qwen/Gemma) and goal (search/free-view) as factors. The main effect of model type was significant, $F(1, 2325) = 17.76$, $p < .001$, $\eta_G^2 = .001$. The main effect of goal was also significant, $F(1, 2325) = 235.52$, $p < .001$, $\eta_G^2 = .034$. There was a significant interaction between model type and goal, $F(1, 2325) = 37.35$, $p < .001$, $\eta_G^2 = .002$. Despite this significant interaction, Figure 8 shows that fixation-order alignment in the search task is near zero for both models. Indeed, planned comparisons show that Qwen had stronger fixation-order alignment in the free-view task than in the search task (free-view mean: 0.008, search mean: ≈ 0; $t(2325) = 16.03$, $p < .001$, 95% CI [0.007, 0.009], $d = 0.47$). Similarly, Gemma also had stronger fixation-order alignment in the free-view task than in the search task (free-view mean: .007, search mean: .003; $t(2325) = 9.93$, $p < .001$, 95% CI [0.004, 0.006], $d = 0.29$). A supplemental analysis (see the Supplementary Materials) suggests that the null results in the search condition are partially due to the inherent center bias of human observers. Specifically, human observers tended to start viewing from the image center regardless of the goal, whereas VLM predictions did not show a center bias under the search condition.

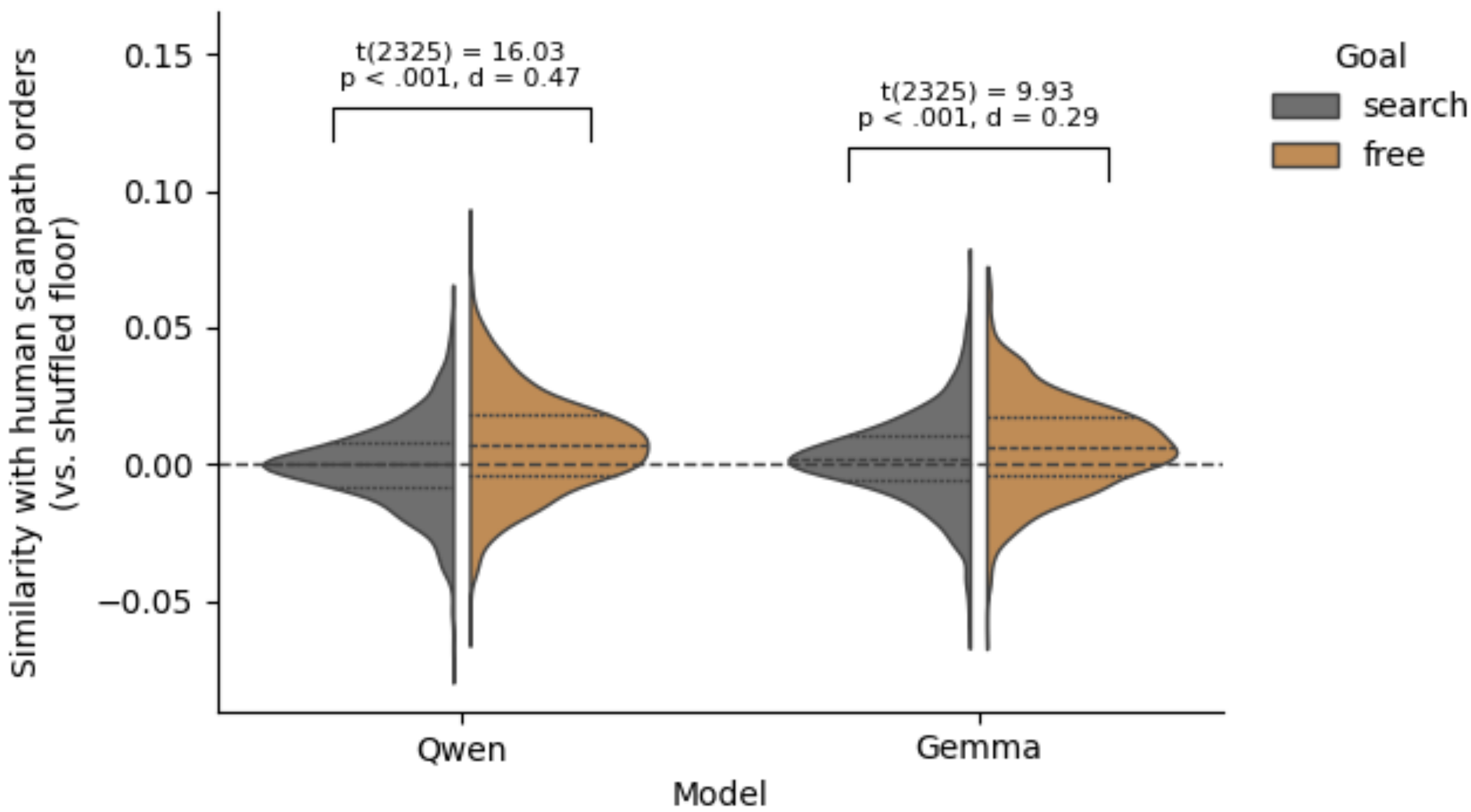

**Figure 8. Scanpath similarity between model-emitted points and human fixations under different task goals.** The split violin plots show smoothed distributions of similarity scores between model-emitted points and human scanpaths, above a shuffled floor to isolate similarity in fixation order. The horizontal dashed lines inside the distributions denote quartiles. The *t*-statistics and Cohen's *d*s are absolute values.

## Discussion

We conducted a series of tests to examine whether large VLMs could predict human goal-directed attention in a zero-shot manner without direct gaze training. We show that VLM predictions align more closely with human fixations when models and humans operate under similar task goals. This crossover effect indicates that VLMs flexibly adjust their predictions in a human-like, goal-directed manner. The target-absent results ruled out that this ability can be explained away by VLMs' visual grounding abilities, as there was no target for VLMs to localize in the first place. Indeed, an analysis of Qwen's thinking traces suggests that the model appears to invoke scene-semantic knowledge (e.g., a bottle should be near the table) on target-absent images. In contrast, during free viewing, the model's reasoning appears to be more rooted in visual prominence (e.g., "The pineapple pie is very eye-catching"). Although model-thinking traces are not necessarily a faithful representation of the underlying computation, they provide converging evidence that VLMs adjust their behavior in a human-like, goal-directed manner in accordance with task goals.

The goal here is to examine goal-directed alignment rather than optimizing model predictive performance. Nevertheless, it is still useful to place these correlations in context. When the model and human goals matched, Qwen and Gemma reached correlations of .73 and .70 on target-present search, .45 and .40 on target-absent search, and .42–.48 on free viewing. For comparison, DeepGaze II [25], a deep saliency model trained on human fixation data, achieved correlations of .24 for target-present search, .30 for target-absent search, and .46 for free viewing on the same dataset [22]. A supervised deep-learning model trained directly on human fixations from the current dataset achieved correlations of .68, .54, and .59, respectively, on the same dataset [22]. Though a more systematic comparison is needed, these results suggest that zero-shot VLM predictions are competitive with models specifically trained to predict human fixations.

The current results connect to a long line of work showing that human attention in real-world scenes is guided by scene semantics [2–6]. Previous studies have typically demonstrated this guidance through controlled manipulations of scene content [5] or extensive human ratings of local meaningfulness [26]. Our results suggest that broad multimodal training may endow VLMs with functionally similar scene-semantic knowledge, enabling them to produce spatial priority predictions that align with human fixations. Furthermore, Qwen's thinking traces explicitly refer to task-relevant objects and likely locations across thousands of scenes. These traces provide human-readable descriptions of the model's scene priors that could help generate testable hypotheses about human semantic guidance beyond what is feasible with conventional approaches.

More generally, the current results may inform how task goals shape information processing. The VLMs examined here do not contain a module explicitly dedicated to exerting control over outputs. Instead, the supplied goal is represented by ordinary text tokens and processed jointly with visual tokens, producing human-aligned, goal-directed behaviors. This property arises within a shared computational system in which supplied goals interact with visual representations through learned associations. These findings support the computational possibility that human cognitive control may likewise stem from general learning and context-dependent retrieval processes, without assuming a dedicated control unit [27,28].

Because Qwen and Gemma do not fully disclose their training data, we cannot determine whether they encountered the images used here during training. Nevertheless, several features of our design make simple image memorization an unlikely explanation for the observed results. Neither model received task-specific training on the fixation data, and mere exposure to an image would not explain why different task instructions produce distinct priority maps that selectively align with human fixations under the corresponding goals. The target-absent condition provides a stronger test: because the target does not appear in the image, successful prediction cannot be reduced to localizing a visible target, forcing the model to identify locations where the target would plausibly occur. Although these findings cannot rule out all effects of training-data exposure, the goal-dependent alignment observed here is more plausibly explained by the models' general scene-semantic knowledge.

It is also important to emphasize that behavioral alignment does not imply mechanistic equivalence between VLMs and humans. VLMs lack an oculomotor system and do not generate predictions through a sequence of fixations. Relatedly, an interesting observation is that, when shown target-absent scenes, Qwen often first concluded that the target was absent and then enumerated locations where it would plausibly occur (see the Supplementary Materials for a complete thinking trace). This pattern suggests that their predictions reflect holistic scene-semantic inference rather than a human-like sequence of search decisions, which may explain why the VLMs failed to predict fixation order during visual search. Consistent with this distinction, human observers exhibited a strong center bias across task goals, whereas the VLMs showed a center bias during the free-viewing memory task but not during search. The models' search predictions, therefore, appear less constrained by the oculomotor and experimental factors that shape human fixation orders. These divergences echo recent evidence that large language models align less closely with humans on sensorimotor than on non-sensorimotor dimensions [29].

Overall, the current results show that state-of-the-art, off-the-shelf vision-language models exhibit human-aligned patterns of goal-directed spatial prioritization without direct training on gaze data. These models, therefore, offer a promising foundation for predicting human attention in specific contexts. In driving, for example, it is critical to anticipate where potential hazards may emerge. The finding that general-purpose VLMs predict human fixations even when a search target is absent suggests that their scene-semantic knowledge could help automated driving systems prioritize regions where hazards may appear. Although realizing this application will require validation using dynamic visual input, the emergence of human-aligned, goal-directed prioritization in large VLMs points to broad potential in contexts that benefit from predicting where people would look.

# Methods

This research collects data from open-weight vision-language models and uses publicly available human participant data. The original studies indicate that the procedures used to collect human participant data were approved by the institutional review board at Stony Brook University [21,22].

## Models

We tested two open-weight, state-of-the-art vision-language models: Qwen3-VL-32B-Thinking [19] (weights released 10/21/2025) and Gemma-4-26B-A4B-it [20] (weights released 4/2/2026). Model weights were downloaded from HuggingFace without modification and hosted on the University's high-performance clusters.

Qwen3-VL-32B-Thinking is a dense 32B-parameter model pretrained with trillions of tokens of multimodal data, including image captions, knowledge collections, optical character recognition data, documents, code, and videos [19]. The model was then refined with supervised fine-tuning, distillation, and reinforcement learning to improve its instruction-following, reasoning, and alignment with human preferences. The model natively supports point-based grounding by emitting an (x, y) coordinate. The Thinking variant also emits an explicit reasoning trace before answering, which is marked by <think>...</think> tags. At inference, we set its temperature to 0.7, top-p to 0.8, top-k to 20, and max tokens to 6144 in a 10240-token context.

Gemma-4-26B-A4B-it is a mixture-of-experts model with 26B total parameters, and approximately 4B active per token. It was pretrained with large-scale data from a wide range of domains and modalities, including web documents, code, and images [20]. The model was then refined with knowledge distillation and reinforcement learning to improve its math, coding, reasoning, instruction-following, and multilingual abilities. At inference, we set its temperature to 0.7, top-p to 0.8, top-k to 20, minimum tokens to 64, and maximum tokens to 420 in an 8192-token context.

## Prompts

For both models, the following prompt was used for the search task:

*"You are a helpful assistant. You are searching for a {target} in this image. If a {target} is clearly visible, point to the {target}; otherwise, point out all locations in this image you would examine before you can decide whether a {target} is present. Rank the locations from most to least likely. Give one location per line in exactly this format:*
*REGION: <a short noun phrase for the place> | POINT: (x, y)*
*where x and y are integers from 0 to 1000, x left-to-right and y top-to-bottom."*

The following prompt was used for the free-view task:

*"You are a helpful assistant. You are viewing this image to remember it for a later memory test. Point out the locations in this image you would naturally look at while studying it. Rank the locations from most to least likely to draw your attention. Give one location per line in exactly this format:*
*REGION: <a short noun phrase for the place> | POINT: (x, y)*
*where x and y are integers from 0 to 1000, x left-to-right and y top-to-bottom."*

The two prompts share the same response format and differ only in their goals. The wording of the memorization prompt in the free-view condition deliberately matches the COCO-FreeView instruction (viewing in anticipation of a memory test).

## Priority map from model output

We asked the model to generate 20 answers separately for each image in response to a given prompt. We then parsed the model's answer to extract the emitted (x, y) points and retained the top five points per response. Two reasons motivated the choice of the top five. First, it guarded against the model exhaustively producing points under the free-view condition. Second, five equated Qwen and Gemma and matched human behavior: it is the median number of fixations in human target-absent search scanpaths on COCO-Search18.

From the emitted points, we constructed a two-dimensional frequency matrix at ¼ of the original image resolution. We then applied a low-pass Gaussian filter with a cutoff frequency of 6 cycles/image, and rescaled values to 0 and 1.

## Priority map from attention weights

We passed the same prompt to the model in a single forward pass and captured attention weights as the model processed the prompt. For the search prompt, we captured attention weights from the target's first mention (i.e., in the sentence "You are searching for a {target} in this image.") to the image tokens, at every layer and attention head within a pre-defined layer band. For Qwen, this band spanned layers 26–44 of its 64 decoder layers (the middle 30%). For Gemma, this band spanned layers 4–12 of its 30 decoder layers (an earlier 30%). In the same forward pass, we also captured a baseline: attention weights from the rest of the prompt (excluding all mentions of the target) to the image tokens. Within each head, we subtracted that head's baseline attention weights from the target's attention weights. The difference matrix was floored at zero to retain locations where the target word drew higher attention weights than the rest of the prompt. We then combined these per-head difference matrices into a single matrix through a weighted average: heads whose attention weights concentrated on a few image locations were weighted more heavily than heads with diffused locations. The rationale here was to enhance potential goal-related signals. Finally, we applied a few post-processing steps: we clipped values to the 99th percentile, set values at two fixed corners to zero, and rescaled the matrix from zero to one. The two corners correspond to attention-sink positions, which are artificial high-weight locations regardless of image content.

For the memory prompt, we extracted the attention weights from the word "remember" (in the sentence "You are viewing this image to remember it for a later memory test") to the image tokens, at a single frozen layer (layer 24 for Qwen and layer 29 for Gemma). The baseline weights were from the rest of the prompt's tokens to the image tokens. All subsequent steps matched the search prompt.

From the post-processed attention weights, we constructed a priority map at ¼ of the original image resolution, with points weighted by the attention weights at that location. We then applied a low-pass Gaussian filter with a cutoff frequency of 6 cycles/image, and rescaled values to 0-1.

## Priority map from human fixations

Human fixation data came from two large-scale, open-access datasets: COCO-Search18 [21] and COCO-FreeView [22]. In COCO-Search18, ten participants searched for each of 18 target-object categories in 6,202 real-world images with eye movements recorded. Search targets included common daily objects, such as a clock, a bottle, a sink, a TV, and an oven. Each trial began with a center fixation dot, followed by the image, which stayed on until participants made a manual target-present/absent response. The number of target-present and target-absent images was equal.

The search images were selected to meet several criteria (no humans/animals, single target object, no extreme size, etc.). Importantly, for the current study, target-absent images were selected to ensure the target matches the scene context (e.g., searching for a sink in the kitchen), so observers cannot indicate target absence based on global context alone.

COCO-FreeView contained the same images as COCO-Search18. A separate group of ten participants were instructed to view the images in preparation for a memory test. Each image was presented for 5 seconds.

For both datasets, only the training and validation sets are publicly available. We reserved 54 images from the training set for pilot testing purposes (see Section "Pilot testing"). The rest of the images (train + validation) were used in the main analyses, which included 4,887 images (2474 target-present, 2413 target-absent) with 166,868 search fixations and 612,346 free-view fixations.

To construct human priority maps, we pooled fixations from correct trials in the search condition and all trials from the free-view condition. The first fixation (center-start) of each trial was excluded. Subsequent steps followed the procedure for model-emitted points: we created a frequency matrix from fixation coordinates at ¼ image resolution, applied a low-pass Gaussian filter with a cutoff frequency of 6 cycles/image, and rescaled values to 0-1. Fixation durations were not used to create human priority maps.

## Evaluation metric

We use Pearson correlation as our evaluation metric, given that it is easy to interpret and balances false positives and false negatives [30]. Following convention, the maps were first normalized to have zero mean and unit variance, then reshaped to one dimension to compute the Pearson correlation.

## Thinking trace analysis

The thinking trace analysis was conducted for target-absent images. Qwen thinking traces were extracted based on the <think>...</think> tags. From raw traces, we extracted sentences that named concrete scene objects (e.g., "The area near the door or the wall"). Target words were removed from the sentence (e.g., “Maybe the is on the floor”, with the target “potted plant” removed). In addition, sentences that only mentioned abstract concepts (e.g., “Let’s examine this image/area/region.”) and those with fewer than 3 words after cleaning were removed. After cleaning, each picture had an average of 621 sentences in the search condition and 603 sentences in the free-view condition, across all model-generated responses.

The semantic anchor for the search condition was the target word (e.g., “bottle”). The semantic anchor for the free-view condition was a collection of seven terms: *salient*, *conspicuous*, *prominent*, *striking*, *noticeable*, *dominant*, and *stands out*. We embedded each cleaned sentence into a 384-dimensional vector using a lightweight sentence transformer model, all-MiniLM-L6-v2 [23]. The semantic anchors were also embedded, with the seven terms in the free-view condition averaged to obtain a single embedding vector. Cosine similarity was used to quantify the similarity between each cleaned sentence and the two semantic anchors. We computed the mean cosine similarity across all sentences to indicate how well the model’s thinking aligns with the target or visual prominence.

## Fixation order analysis

The fixation order analysis was also conducted for target-absent images only, as there were too few human fixations (2-3 fixations on average) on target-present images to meaningfully examine order effects. For each model response, we selected the top five emitted points as model scanpaths. Human scanpaths came from each correct trial’s fixation sequence, with the initial center-start fixation removed and truncated to the first five fixations. Scanpath similarity was computed from a Python re-implementation of an existing distance-based edit distance metric implemented in the R package *scanpath* [24]. Here, we used unit fixation durations and assumed a 30-degree visual angle of the image width. We also converted the original dissimilarity score to a similarity score by normalizing it by scanpath length and subtracting from one. This gives us a score from 0 to 1, with a higher score indicating greater similarity. All other parameters adopted the default settings of the original code.

For each image, pairwise similarity scores were computed for all possible model-human combinations, and the mean was used to indicate the overall similarity between model and human scanpaths on that picture. To isolate the effect of order beyond spatial distribution, we constructed a baseline by averaging the similarity scores from five random shuffles of the model scanpath while holding overall locations fixed. The alignment in fixation order is thus the model similarity minus the baseline.

## Pilot testing

Using 54 images from the training set, we tested several parameter choices and then froze them. Specifically, the attention-layer bands for the search condition were selected to maximize their correlations with the corresponding human fixation density map (Qwen: layers 26–44; Gemma: layers 4–12). For the free-view condition, a single layer was selected (Qwen: layer 24; Gemma: layer 29), as no broadband layers aligned with free-view human fixations. Using the same images, we also debugged several key aspects of the analysis to ensure they work as intended: (1) the model prompts produced sensible output that can be parsed for downstream analyses, (2) the priority maps looked sensible, (3) the thinking trace analysis pipeline worked, and (4) jobs can be submitted to the University’s high-performance clusters.

## Statistical analyses

All primary analyses were performed using a 2*2 repeated-measures ANOVA, assuming image as the independent unit of analysis. The exact number of images slightly varies across analyses because some model outputs were unparseable. To evaluate the alignment between model priority maps and human priority maps, we conducted a 2*2 repeated-measures ANOVA with model goal (search vs. free-view) and human goal (search vs. free-view) as factors, separately for target-present and target-absent images. The dependent variable was the Pearson correlation between the model maps and human maps. The thinking trace analysis used a 2*2 repeated-measures ANOVA with model goal (search vs. free-view) and semantic content (target vs. prominence) as factors. The dependent variable was the mean cosine similarity per image. For the fixation order analysis, we conducted repeated-measures ANOVA with model type (Qwen vs. Gemma) and goal (search vs. free-view) as factors. The dependent variable was baseline-controlled scanpath similarity. For ANOVAs, we report main effects and interactions, using generalized eta-squared ($\eta_G^2$) as effect size. Planned follow-up comparisons used two-tailed paired $t$-tests and Cohen's $d$ as effect size. The scene-level reorganization analysis used two-tailed one-sample $t$-tests and Cohen's $d$ as effect size. We report 95% confidence intervals of mean differences for all $t$-tests. All analyses used a significance threshold of $\alpha = .05$.

## Data availability

Model output data necessary to reproduce results reported in the manuscript are available at https://zenodo.org/records/22151952. Human fixation data and images from COCO-Search18 and COCO-FreeView [21,22] are available at https://sites.google.com/view/cocosearch/home.

## Code availability

The analysis code necessary to reproduce results reported in the manuscript is available at https://github.com/HanZhang-psych/VLM-goal-directed-attention. Model weights for the VLMs tested in the manuscript can be downloaded at https://huggingface.co/Qwen/Qwen3-VL-32B-Thinking and https://huggingface.co/google/gemma-4-26B-A4B-it.

# Supplementary Materials

## Priority maps from attention weights

To test whether such goal alignment is intrinsic to the model's computational processes, we directly extracted attention weights from Qwen's and Gemma's decoder layers and constructed priority maps based on these extracted weights.

### Target-present images

We conducted a 2*2 within-image repeated-measures ANOVA with model goal (search/free) and human goal (search/free) as factors. For Qwen, there was a main effect of model goal, $F(1, 2473) = 18.39$, $p < .001$, $\eta_G^2 = .001$, a main effect of human goal $F(1, 2473) = 421.71$, $p < .001$, $\eta_G^2 = .04$, and importantly, a significant interaction between model goal and human goal, $F(1, 2473) = 6110.37$, $p < .001$, $\eta_G^2 = .28$. Planned comparisons show a crossover effect: Qwen search predictions aligned significantly better with human search fixations than with human free-view fixations (average $r_{search\text{-}search} = .53$, average $r_{search\text{-}free} = .13$; $t(2473) = 62.54$, $p < .001$, 95% CI [0.389, 0.414], $d = 1.68$). Conversely, Qwen free-view predictions aligned significantly better with human free-view fixations than with human search fixations (average $r_{free\text{-}free} = .41$, average $r_{free\text{-}search} = .22$; $t(2473) = 31.36$, $p < .001$, 95% CI [0.183, 0.207], $d = 0.80$).

For Gemma, there was a main effect of model goal, $F(1, 2473) = 5519.36$, $p < .001$, $\eta_G^2 = .37$, and a main effect of human goal, $F(1, 2473) = 5395.84$, $p < .001$, $\eta_G^2 = .247$. Importantly, there was also a significant interaction between model goal and human goal, $F(1, 2473) = 9004.39$, $p < .001$, $\eta_G^2 = .472$. Gemma search predictions aligned significantly better with human search fixations than with human free-view fixations (average $r_{search\text{-}search} = .63$, average $r_{search\text{-}free} = .07$; $t(2473) = 120.47$, $p < .001$, 95% CI [0.551, 0.569], $d = 3.06$). Similarly, Gemma free-view predictions aligned significantly better with human free-view fixations than with human search fixations (average $r_{free\text{-}free} = .13$, average $r_{free\text{-}search} = -.004$; $t(2473) = 29.43$, $p < .001$, 95% CI [0.129, 0.147], $d = 0.74$).

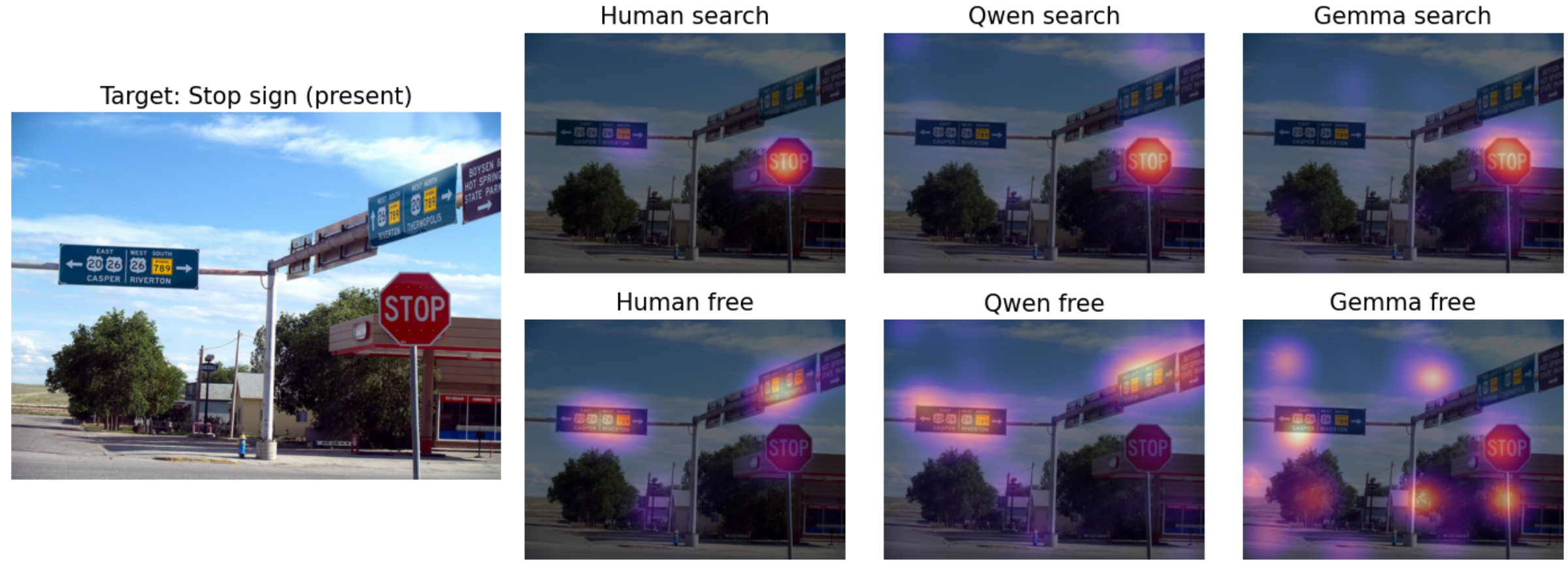

**Figure S1. Human and VLM priority maps for a target-present image.** VLM priority maps are constructed from attention weights extracted from model decoder layers.

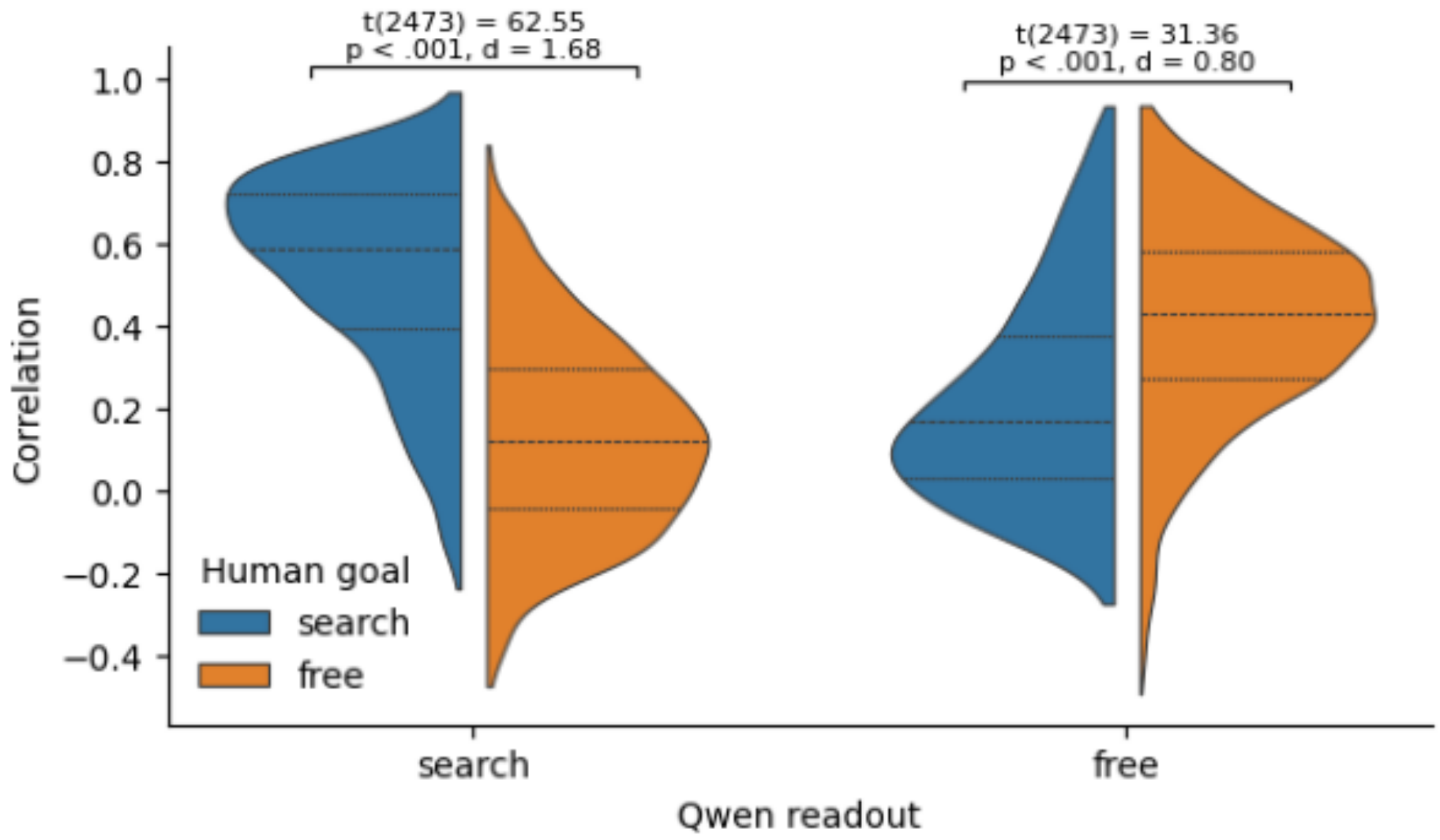

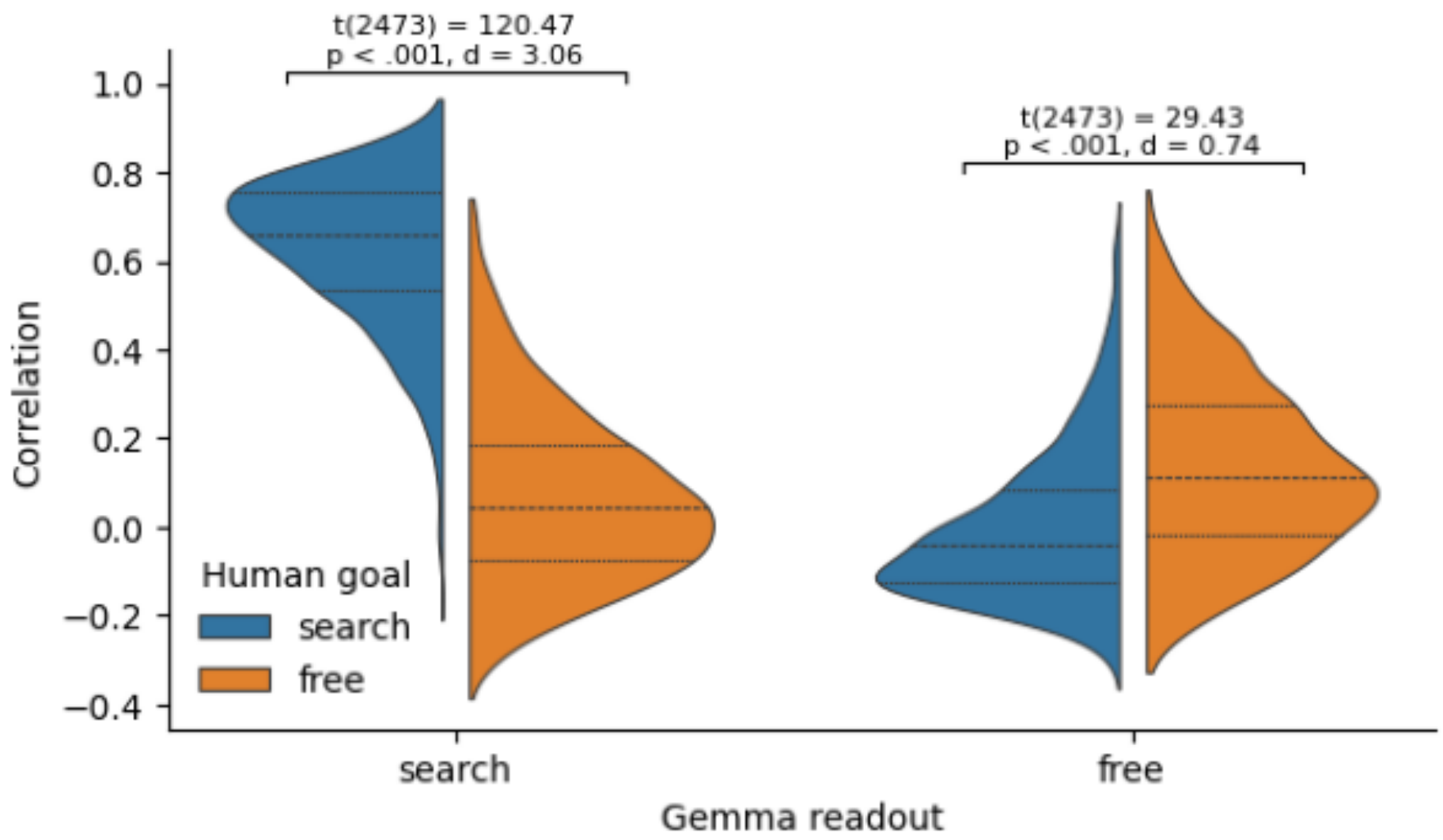

**Figure S2. Goal alignment between VLMs and human observers (target present).** The upper panel shows results for Qwen, and the bottom panel shows results for Gemma. VLM priority maps were constructed directly from model decoder layers.

#### Target-absent images

We again conducted a 2*2 repeated-measures ANOVA with model goal (search/free) and human goal (search/free) as factors. For Qwen, there was a main effect of model goal, $F(1, 2412) = 1825.18$, $p < .001$, $\eta_G^2 = .120$, and a main effect of human goal, $F(1, 2412) = 67.82$, $p < .001$, $\eta_G^2 = .005$. Importantly, the interaction between model goal and human goal was also significant, $F(1, 2412) = 1317.71$, $p < .001$, $\eta_G^2 = .037$. Planned comparisons show that Qwen search predictions aligned significantly better with human search fixations than with human free-view fixations (average $r_{search\text{-}search} = .21$, average $r_{search\text{-}free} = .15$, $t(2412) = 11.67$, $p < .001$, 95% CI [0.049, 0.069], $d = 0.24$). Conversely, Qwen free-view predictions aligned significantly better with human free-view fixations than with human search fixations (average $r_{free\text{-}free} = .42$, average $r_{free\text{-}search} = .29$, $t(2412) = 26.40$, $p < .001$, 95% CI [0.120, 0.140], $d = 0.57$).

For Gemma, there was a main effect of model goal, $F(1, 2412) = 49.00$, $p < .001$, $\eta_G{}^2 = .007$, a main effect of human goal, $F(1, 2412) = 7.54$, $p < .001$, $\eta_G{}^2 < .001$, and a significant interaction between model goal and human goal, $F(1, 2412) = 2022.73$, $p < .001$, $\eta_G{}^2 = .100$. Planned comparisons show that Gemma search predictions aligned significantly better with human search fixations than with human free-view fixations (average $r_{search\text{-}search} = .15$, average $r_{search\text{-}free} = .03$, $t(2412) = 29.55$, $p < .001$, 95% CI [0.116, 0.132], $d = 0.57$). Conversely, Gemma free-view predictions aligned significantly better with human free-view fixations than with human search fixations (average $r_{free\text{-}free} = .13$, average $r_{free\text{-}search} = -0.01$, $t(2412) = 35.69$, $p < .001$, 95% CI [0.132, 0.147], $d = 0.75$).

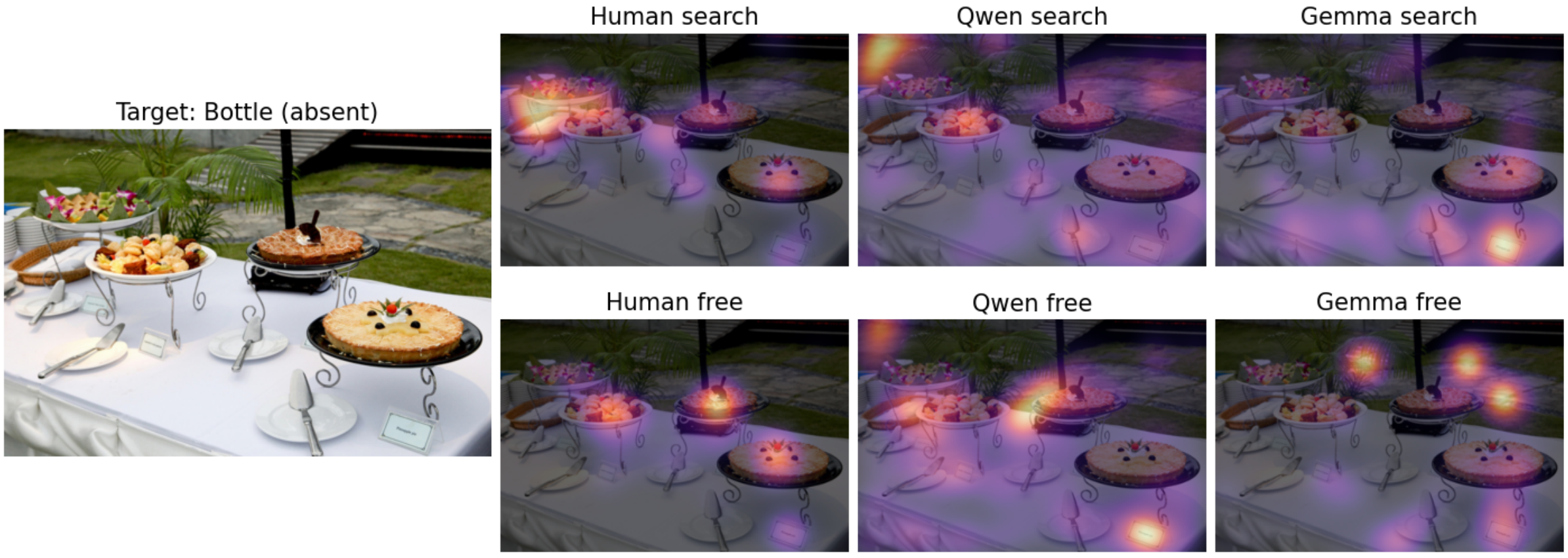

**Figure S3. Human and VLM priority maps for a target-absent image.** VLM priority maps are constructed from attention weights extracted from model decoder layers.

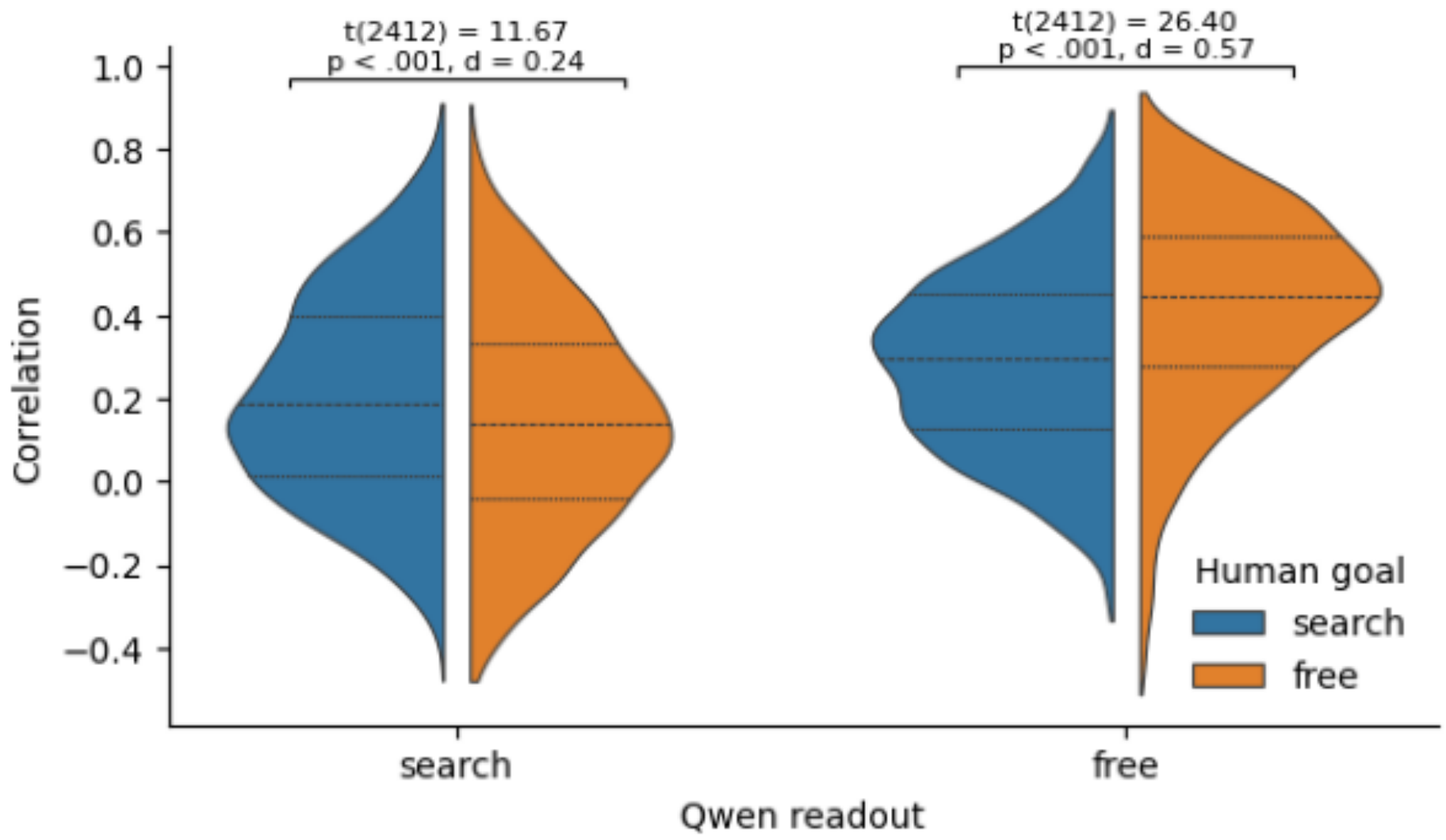

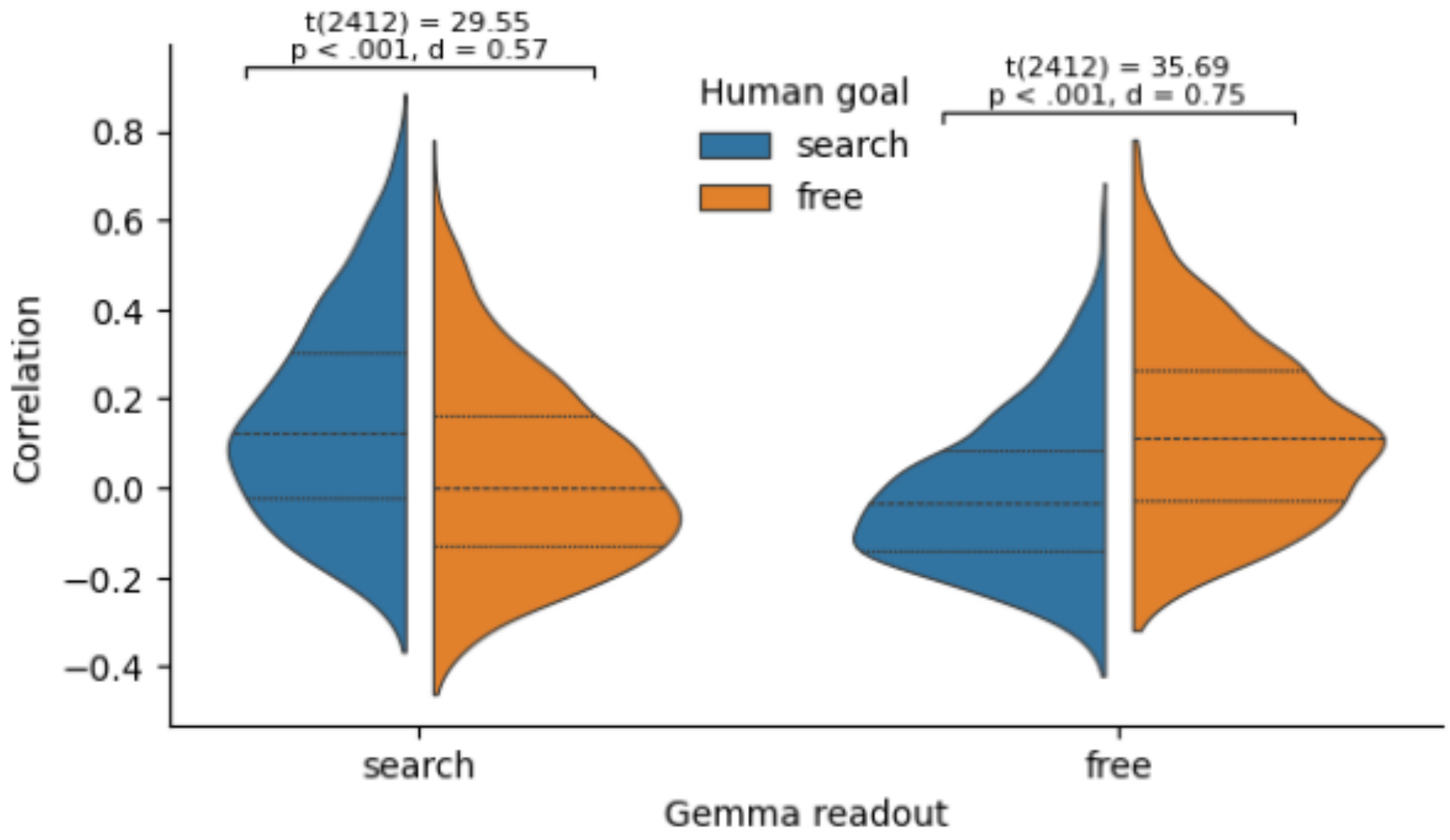

**Figure S4. Goal alignment between VLMs and human observers (target absent).** The upper panel shows results for Qwen, and the bottom panel shows results for Gemma. VLM priority maps were constructed directly from model decoder layers.

## Mean fidelity values for each target object category

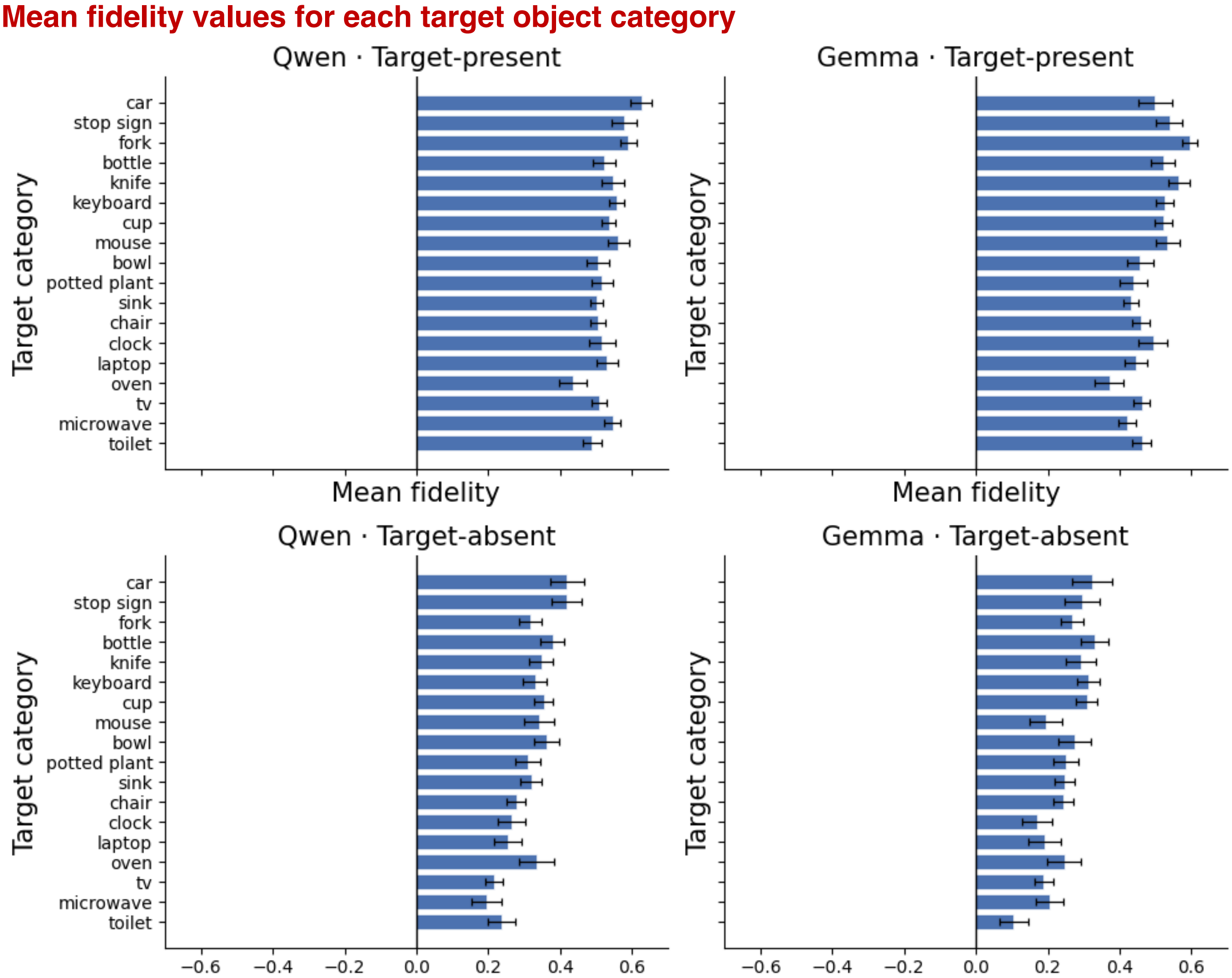

**Figure S5. Mean fidelity values for each target object category with 95% CIs.** A positive value means VLMs align with humans on which scene regions they prioritize and deprioritize when the task goal changes.

## Semantic dissociation for each search object category

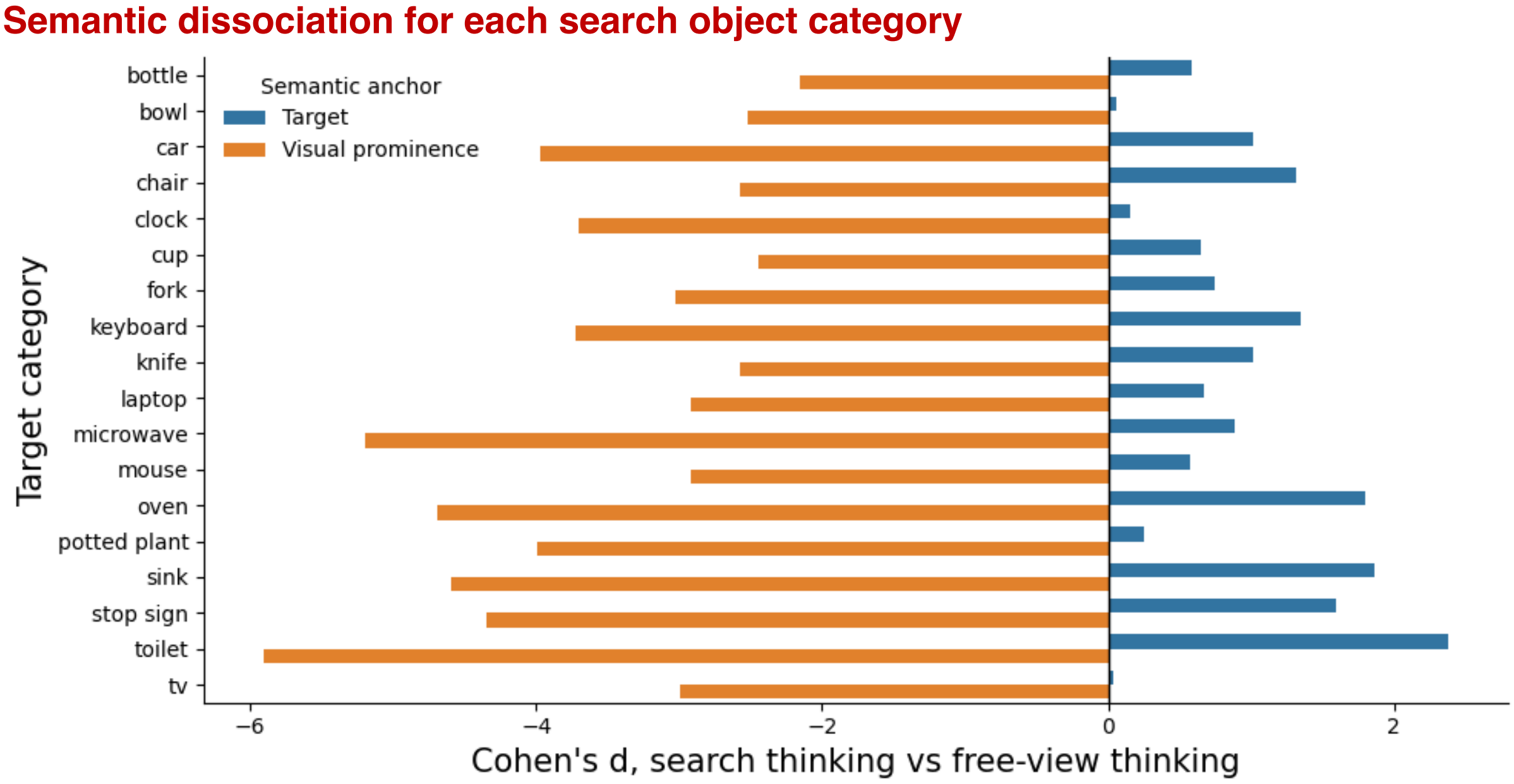

**Figure S6. Semantic dissociation for each search object category.** Values are Cohen's d for search minus free-view thinking for each semantic anchor. Positive values indicate that search thinking is more target-aligned

than free-view thinking. Negative values indicate that free-view thinking is more aligned with visual prominence than search thinking.

## An example Qwen thinking trace for the Figure 3 image

*Got it, let's look at the image. The task is to find a bottle or check where a bottle might be. First, check all areas. The image shows a buffet table with cakes, plates, utensils, and background elements. Let's list possible places:*
*The area behind the table, maybe near the plants or the stone path. But the background has grass, plants, and a structure. Wait, the table has various items. Let's check the left side: there are stacked plates, a wicker basket. The background has a black pole (maybe part of an umbrella), but no bottle. The main table has food items, no bottles. So possible places to check:*
*The area near the stacked plates on the left: maybe a bottle is hidden there. The stacked plates are on the left, so check around them.*
*The background, like the grassy area or near the stone path. But that's less likely.*
*The area behind the cakes, maybe near the black pole. But the pole is a stand, not a bottle.*
*Wait, the image doesn't show any bottle. So we need to list locations to check. The most likely places would be where bottles are commonly placed, like near the table, maybe behind the food displays or in the background.*
*Let's list:*
*The area behind the stacked white plates on the left (since sometimes bottles are near serving areas). So check there.*
*The background near the green plants and the stone path, maybe a bottle is there.*
*The area around the black pole (the umbrella stand), maybe a bottle is placed there.*
*But need to rank from most to least likely. Let's think: in a buffet setup, bottles (like for drinks) might be near the serving area. The left side has stacked plates, so maybe a bottle is there. Let's check the left side first.*
*So:*
*REGION: area behind stacked plates | POINT: (50, 320)*
*REGION: area near the black pole | POINT: (585, 200)*
*REGION: background grassy area | POINT: (750, 250)*
*Wait, but the image shows the table with food, and no bottles are visible. So the locations to check are where a bottle might be hidden. Let's check the left side where the plates are stacked. The stacked plates are on the left, so the space behind them (towards the back of the table) might have a bottle. Also, the area near the wicker basket. Let's see:*
*The first possible spot is the left side, behind the stacked plates. The plates are on the left, so the area behind them (x around 50, y around 300) is a place to check.*
*Next, the area near the black pole (the umbrella stand) which is in the middle background. Maybe a bottle is there.*
*Then, the background grassy area, but that's less likely.*
*So the order would be:*
*The area behind the stacked white plates on the left (since that's a common place for serving items, maybe a bottle for drinks)*
*The area near the black pole (umbrella stand) in the background*
*The background grassy area (least likely, as it's not a serving area)*
*Now, convert to the required format.*

## Center bias in humans and VLMs

We computed normalized eccentricity for each model-emitted point and human fixation relative to the image center, with lower values indicating fixations closer to the center. Figure S7 shows mean eccentricity as a function of fixation index. Human observers showed a clear center bias under both search and free-view goals, although search fixations were somewhat more eccentric than free-view fixations. In contrast, VLM predictions showed a center-biased pattern only under the free-view goal. To formally test this, we conducted a 2*2 repeated-measures ANOVA on mean first-fixation eccentricity, with source (model/human) and goal (search/free-view) as within-image factors. There was a significant main effect of source, $F(1, 2412) = 2617.39$, $p < .001$, $\eta_G^2 = .160$, a significant main effect of goal, $F(1, 2412) = 1148.14$, $p < .001$, $\eta_G^2 = .142$, and a significant source × goal interaction, $F(1, 2412) = 275.80$, $p < .001$, $\eta_G^2 = .021$.

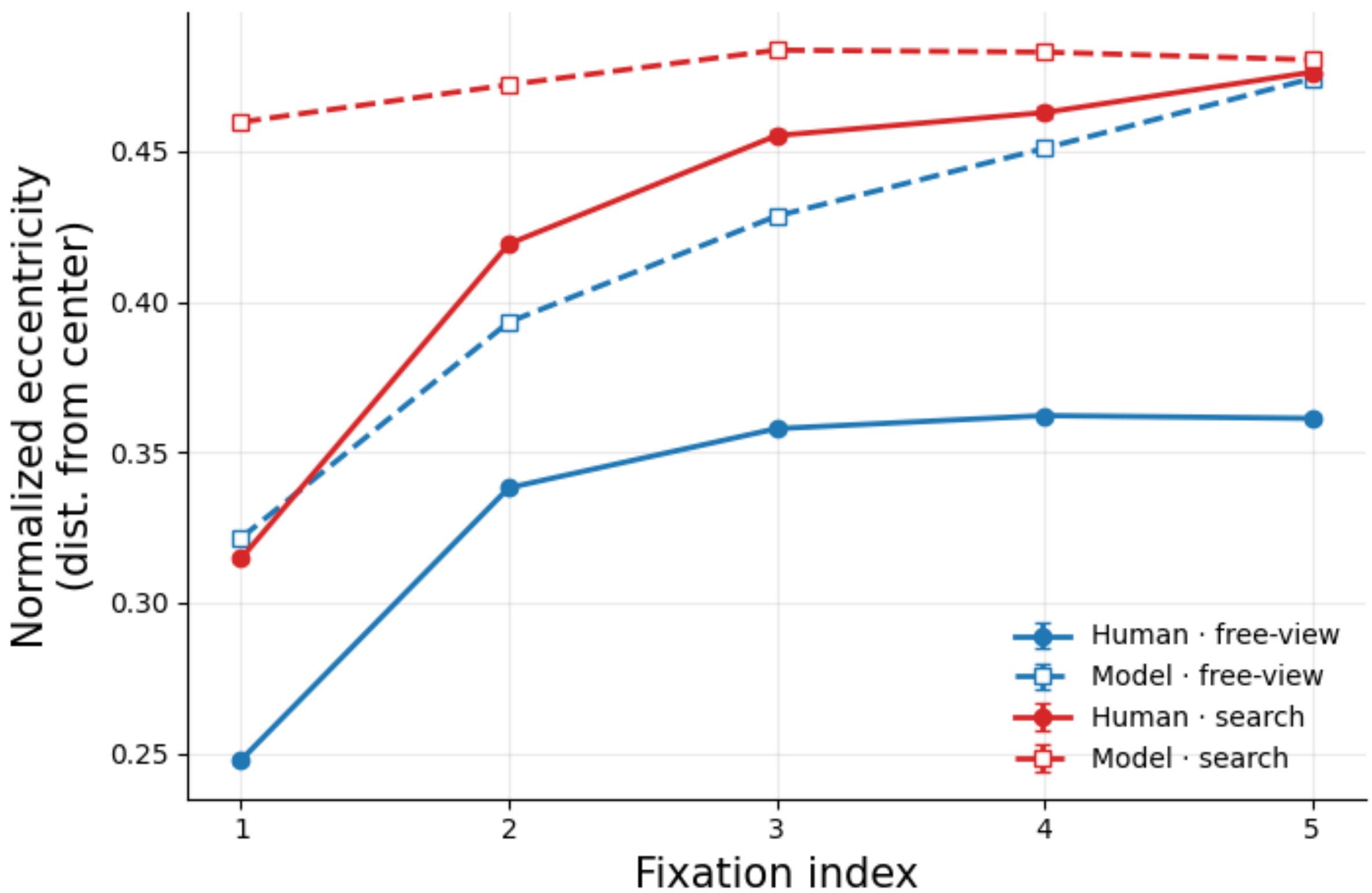

**Figure S7. Center bias in humans and VLMs (target-absent images, Qwen and Gemma combined).**